\pdfoutput=1

\documentclass[10pt,twocolumn,letterpaper]{article}

\usepackage{wacv}%

\usepackage{array}%
\usepackage{float}
\usepackage{algorithm}
\usepackage{algpseudocode}
\usepackage{placeins}
\usepackage{tocloft}%
\usepackage{dblfloatfix}%
\usepackage{pgfplots}
\pgfplotsset{compat=1.17}

\definecolor{surveyOurs}{RGB}{62,110,196}
\definecolor{surveyAnimateX}{RGB}{232,151,58}
\definecolor{surveyBase}{RGB}{76,150,80}
\definecolor{surveyMoore}{RGB}{176,84,92}

\usepackage{adjustbox}

\definecolor{wacvblue}{rgb}{0.21,0.49,0.74}
\usepackage[pagebackref,breaklinks,colorlinks,allcolors=wacvblue]{hyperref}

\def\wacvPaperID{1}%
\def\confName{WACV}
\def\confYear{2027}

\title{One Image Is Enough: Self-Generated Supervision for Personalizing Human Video Diffusion at Test Time}

\author{
Sadman Rakib Pinon$^{1,2}$\quad
Adamo Orsini$^{1,2}$\quad
Yanis Falaki$^{1}$\quad
Linlian Jiang$^{1,2}$\\
Jacob Kingi$^{1,2}$\quad
Yang Wang$^{1,2\dagger}$\quad
Xinxin Zuo$^{1\dagger}$\\[2pt]
$^{1}$Concordia University\qquad
$^{2}$Mila -- Quebec AI Institute\\
{\tt\small sadmanrakib.pinon@mail.concordia.ca}\quad{\small $^\dagger$Corresponding authors}
}

\newcommand{\paperteaser}{%
  \begin{center}
    \vspace{-0.4em}
    \includegraphics[width=0.98\linewidth]{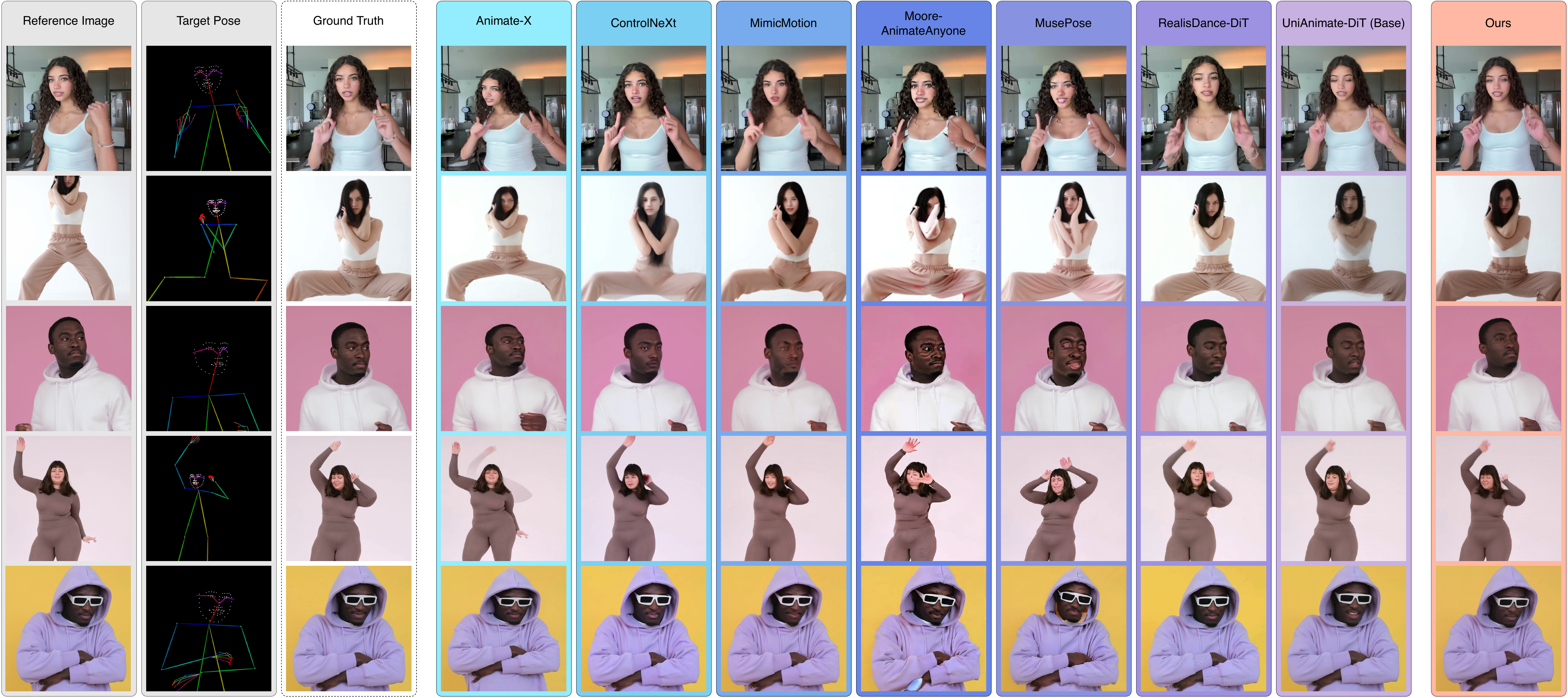}
    \captionsetup{type=figure,hypcap=false}
    \captionof{figure}{\textbf{One image is enough.} From a single reference
    image and a driving pose sequence, our method personalizes a pose-driven
    video diffusion model to the subject at test time, with no subject video,
    ground truth, or labels. Identity preservation and video quality improve
    together. Columns show the reference image and target pose, the ground
    truth, seven pose-driven baselines, and our adapted model.}
    \label{fig:teaser}
    \vspace{0em}
  \end{center}
}

\begin{document}
\AddToHookNext{shipout/background}{\put(\LenToUnit{0.8125in},\LenToUnit{-10.68in}){\parbox[b]{6.875in}{\scriptsize\centering Accepted to the IEEE/CVF Winter Conference on Applications of Computer Vision (WACV) 2027. \copyright~2027 IEEE. Personal use of this material is permitted. Permission from IEEE must be obtained for all other uses, in any current or future media, including reprinting/republishing this material for advertising or promotional purposes, creating new collective works, for resale or redistribution to servers or lists, or reuse of any copyrighted component of this work in other works.}}}

\twocolumn[
  \maketitle
  \vspace{-30pt}
  \paperteaser
]
\addtocontents{toc}{\protect\setcounter{tocdepth}{-1}}
\begin{abstract}
Pose-driven human video diffusion animates a person from a single reference image and a target pose sequence, yet it struggles to preserve identity, rendering the face and garments unfaithfully as the pose departs from it. Personalization is the standard remedy, but its supervision is missing: one image fixes identity at a single pose, while a multi-pose subject video cannot be assumed in practice. Instead, we synthesize the missing supervision. A domain-tuned, pose-conditioned generator produces pseudo-ground-truth frames at selected key poses. An inference-time reference-only sampler steers these candidates toward the subject and keeps the most identity-consistent. A per-subject low-rank adapter then trains on static clips of them, configured from the reference alone, with no subject video, ground truth, or labels. Although lower in fidelity, these synthetic frames supervise more effectively than real ones. Real frames as captured carry motion that erodes the temporal prior, while a static clip conveys appearance alone. On TikTok and UBC Fashion, the adapted model improves the base on every metric, inverts the identity-versus-motion trade-off, and matches or surpasses adaptation on five real frames despite using none. Code and weights are available at \href{https://github.com/SadmanPinon/one-image-is-enough}{\small\texttt{github.com/SadmanPinon/one-image-is-enough}}.
\end{abstract}

\iftoggle{wacvfinal}{}{\vspace*{-12pt}}
\section{Introduction}
\label{sec:intro}
Given a single reference image of a person and a driving pose sequence, pose-driven human video diffusion models aim to animate the reference subject under the target motion. Recent advances in large-scale video diffusion have substantially improved motion realism, temporal coherence, and overall visual fidelity. However, it is still a challenging problem to faithfully preserve the subject’s identity. For example, facial appearance may be generated inconsistently across frames, while fine-grained garment details, text, and logos can be distorted, altered, or hallucinated. 

Rather than relying solely on a larger video diffusion backbone, we focus on personalizing the model to the input subject at test time. A single reference image, however, provides limited adaptation signals: it captures the subject in only one pose and cannot supervise identity preservation across the diverse poses in the driving sequence. We therefore construct pose-diverse pseudo-supervision at inference time.

Specifically, a pose-conditioned, reference-following image diffusion model synthesizes subject-specific images for poses sampled from the driving sequence. More importantly, an inference-time particle-steering procedure, guided by body-similarity and facial-identity rewards, selects high-quality identity-consistent samples without subject videos, target-frame annotations, or human feedback.

Given the pseudo ground truth frames, directly adapting a video diffusion model to static subject images can weaken its motion prior, reducing dynamics and temporal consistency. This identity-and-motion trade-off ~\cite{chefer2024stillmoving,zhao2024motiondirector,li2025personalvideo,li2025magicid} is particularly challenging for unified attention video DiTs ~\cite{wang2025unianimatedit,zhou2025realisdancedit,ding2025mtvcraft}, whose 3D self-attention jointly models appearance and motion without separable appearance-only parameters~\cite{wang2025unianimatedit,zhou2025realisdancedit,ding2025mtvcraft}.
We address this issue by repeating each synthetic image into a static clip and optimizing a lightweight subject-specific LoRA on these appearance-focused clips. This strategy improves identity preservation while retaining the backbone’s motion capability and video quality.

We validate the approach on the TikTok and UBC Fashion benchmarks. The largest gain is on video quality: adaptation substantially lowers FVD where naive personalization is expected to degrade it, so identity and video quality improve together rather than trading off. The same adaptation transfers to in-the-wild videos unseen during training, since the supervision is generated for each test subject rather than learned from a dataset-specific prior. The adaptation restores the face and fine garment detail that the base model distorts as the body turns away from the reference pose. Our main contributions are:
\begin{itemize}
  \item We propose a test-time personalization framework for pose-driven human video diffusion from a single image, requiring no ground truth, no real captured frames, and no human feedback.
  \item A reward-guided generation approach is proposed for synthesizing pseudo-ground-truth images, using body-similarity and facial-identity rewards that measure consistency with the reference image.
  \item Our method achieves the best results on most metrics across benchmarks, surpassing the non-personalized base on every metric and matching or surpassing a model personalized with real captured subject frames.
\end{itemize}

\section{Related Work}
\label{sec:related}

\noindent\textbf{Pose-driven human animation.}
Pose-driven human image animation has converged on the setting we study: a single
reference image is animated according to a target body-pose or motion sequence. Early diffusion models established the
now-standard recipe of reference appearance injection, pose-guided denoising,
and temporal modules for video synthesis~\cite{hu2023animateanyone,
xu2024magicanimate}. Follow-up work makes this recipe practical in open-source
form~\cite{tmelyralab2024musepose,moorethreads2024mooreanimateanyone} and improves
pose confidence and long-video fusion~\cite{zhang2025mimicmotion}. Others inject
SMPL-derived depth, normal, semantic, and skeleton cues~\cite{zhu2024champ},
study efficient control modules~\cite{peng2024controlnext}, and unify reference
appearance and target-pose conditioning in a single video diffusion
branch~\cite{wang2024unianimate}. Recent DiT-based work further shows that
larger video backbones, compact pose encoders, motion tokens, text/background
control, and hard-motion training can substantially improve pose following and
temporal quality~\cite{zhou2025realisdancedit,wang2025unianimatedit,
zhao2025dynamictrl,ding2025mtvcraft,lin2025omnihuman,xu2025hypermotion}.

These works are primarily trained as feed-forward general animators,
and do not adapt the model to the test subject by constructing pose-specific
supervision from the given reference image. Our work therefore focuses on reducing
per-subject identity, clothing, and hard-pose errors after the target motion is
known, rather than introducing another generic pose-control branch.

\noindent\textbf{Personalization and test-time adaptation.}
Diffusion personalization methods show that few-shot subject binding can be
achieved by fine-tuning model weights, learning new tokens, or updating compact
adapter parameters~\cite{ruiz2023dreambooth,gal2023textualinversion,
kumari2023customdiffusion,hu2022lora,han2023svdiff}. Video customization work
highlights a related but sharper problem: static subject reconstruction can
damage the motion prior, reducing dynamics, temporal consistency, or prompt
adherence. Prior video customization methods study one-shot video tuning from
real videos~\cite{wu2023tuneavideo}. Others add subject and motion adapters,
image-prompt conditioning, frozen videos, dynamic denoising weights, reward
supervision, preference pairs, and DiT LoRA tuning~\cite{wei2024dreamvideo,
jiang2024videobooth,zhao2024motiondirector,ma2024magicme,chefer2024stillmoving,
wu2025customcrafter,li2025personalvideo,li2025magicid,ma2025followyourmotion}.
These methods diagnose the same identity-versus-motion tension that appears in
pose-driven animation. Still-Moving~\cite{chefer2024stillmoving} originates the
frozen-clip supervision we adopt, training adapters on still images replicated over
time. There the construction is a liability: it leaves the model producing nearly
static videos unless a Motion Adapter on the temporal-attention layers, trained on
$1{,}600$ real videos, absorbs the effect. Neither remedy transfers, since a unified
3D-attention DiT has no temporal-only layer and we use no real video. The difference
is what the conditioning asks for. A text prompt demands motion the frozen clip does
not show, whereas our pose is held constant across the clip. StableAnimator~\cite{tu2025stableanimator} attacks the
same problem by training an identity adapter into a video diffusion model and
steering denoising with a face reward at inference. The reward there acts on the
delivered video, whereas ours acts on the supervision frames and the animator is
never rewarded. Wan-Dancer~\cite{huang2026wandancer} builds long dance videos on the
same Wan backbone from music rather than a driving pose, and names identity
consistency over long sequences as open future work, which is the failure mode our
per-reference adaptation targets.

In contrast, our input is one still reference image rather than a subject video,
the control signal is an explicit target pose sequence, and we adapt at test time on
pseudo-examples tied to the poses the model must render.

\begin{figure*}[t]
  \centering
  \includegraphics[width=\linewidth]{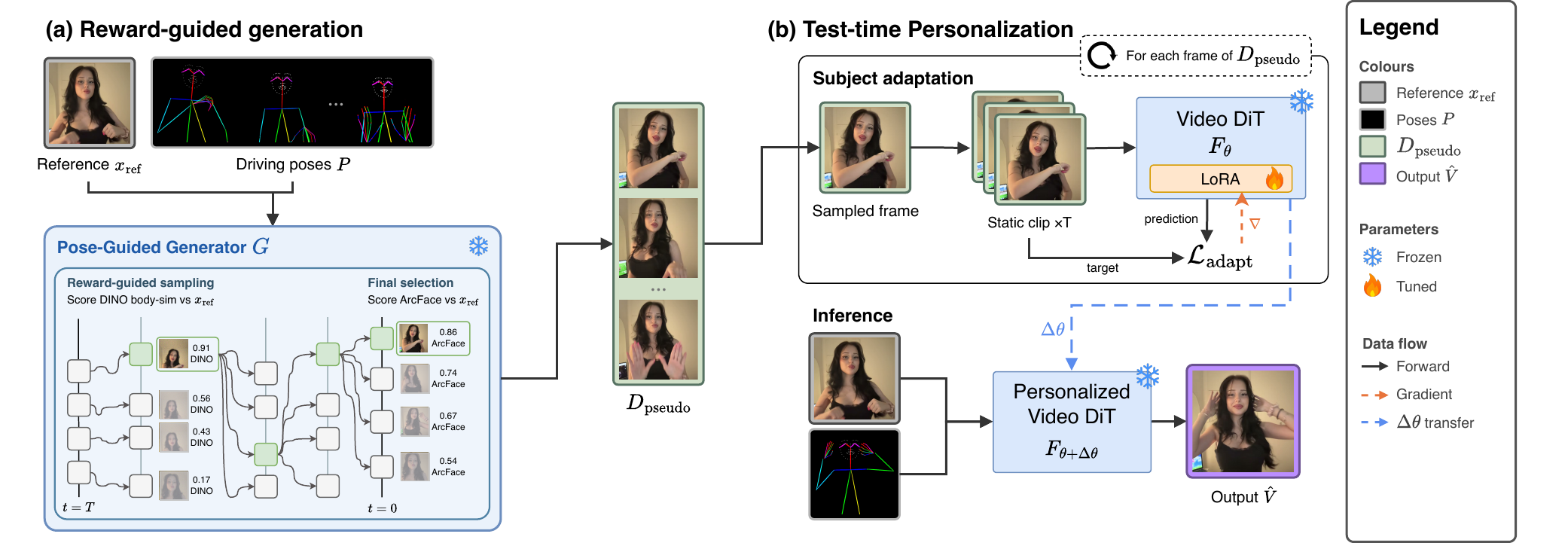}
  \caption{\textbf{Method overview.} From one reference image $x_{\mathrm{ref}}$
  and a driving pose sequence $\mathcal{P}$, a pose-conditioned generator $G$
  synthesizes pose-targeted pseudo-ground-truth frames $\mathcal{D}_{\mathrm{pseudo}}$
  of the subject. \textbf{(a)~Reward-guided generation:} for each key pose, $G$
  draws several candidates as an interacting particle system and resamples them
  in-loop toward a DINO body-similarity reward to $x_{\mathrm{ref}}$, then selects
  the final frame by ArcFace face identity. Every reward is read from the reference
  alone (our reference-only reward-guided sampler). \textbf{(b)~Test-time
  personalization:} a per-subject LoRA $\Delta\theta$ adapts the frozen base
  animator $F_{\theta}$ on static clips of the selected frames. The adapted model
  $F_{\theta+\Delta\theta}$ renders the personalized video $\hat{V}$. No subject
  video, no ground truth, and no human feedback are used.}
  \label{fig:method}
\end{figure*}

\noindent\textbf{Self-generated and synthetic supervision.}
Generated supervision is also well established in diffusion personalization and
view synthesis. Prior work uses generated class images for prior
preservation~\cite{ruiz2023dreambooth}, distills synthetic expert
customizations into amortized personalization models~\cite{chen2023suti}, and
builds synthetic preference pairs for image
personalization~\cite{ayupov2025dreamboothdpo}. Other methods use synthetic
motion, generated views, uncertainty weighting, hallucination maps, or
partial-reference quality measures to avoid poisoning downstream
optimization~\cite{jin2026syntheticmotion,li2026anylift,topaloglu2026oraclegs,
liu2026had,choi2026priqa}. Test-time self-improvement and generated-support-view
pipelines have also appeared for video generation and sparse-view
NVS~\cite{long2026vista,song2026worldforge,neurips2025fewglimpses,
neurips2025dynamicnvs}.

The distinction in our setting is the target and use of the generated supervision.
The single test reference produces candidate human frames at poses sampled from the
driving sequence. Those candidates are filtered for identity and image quality, and
the selected pseudo-pairs supervise per-subject adaptation of the pose-driven video
model.
Thus, the self-generated data are neither generic class regularizers nor
free-view support images, but pose-targeted evidence for the exact
subject-motion pair being requested.

\noindent\textbf{Inference-time guidance and reward-guided sampling.}
A complementary line steers diffusion at inference time without retraining, either
by resampling a population of particles toward a reward~\cite{singhal2025fksteering}
or by searching the noise space with verifiers~\cite{ma2025inferencescaling}.
DINO- and identity-based rewards are standard signals for subject
fidelity~\cite{oquab2024dinov2, deng2019arcface, wei2024personalizedrl}, and best-of-$N$ selection is the simplest
such scheme. Closest to our use, IPRO~\cite{shen2026ipro} optimizes an ArcFace identity reward, but as
\emph{train-time} reinforcement learning of the video model, and IPRO scores
against \emph{ground-truth} video frames.

We instead apply a training-free, reference-only sampler at the pseudo-supervision
stage. A DINO body-similarity reward steers candidate frames in-loop, and an ArcFace
reward selects the final one, anchored entirely to the single reference image, with
no ground truth and no model fine-tuning. Our contribution is this adaptation to the single-reference
video-personalization setting.

\section{Method}
\label{sec:method}

Our goal is to personalize a frozen pose-driven video model to a subject from a
single reference image, with no subject video, ground-truth frames, or labels
available. We therefore synthesize the supervision. From the reference image
$x_{\mathrm{ref}}$ and the driving pose sequence $\mathcal{P}$, a pose-conditioned
generator produces a pseudo-ground-truth frame of the subject at each of a few key
poses sampled from $\mathcal{P}$. Each frame is steered and selected by its identity
match to the reference (\cref{sec:method:pseudo}). A per-subject LoRA then adapts the
base animator on static clips of these frames (\cref{sec:method:tta}). \cref{fig:method}
shows the pipeline, and \cref{alg:testtime} states it in full. We begin with the
preliminaries and problem setup (\cref{sec:method:prelim}).

\subsection{Preliminaries}
\label{sec:method:prelim}

At test time we are given a single reference image $x_{\mathrm{ref}}$ and a driving
pose sequence $\mathcal{P}=\{p_t\}_{t=1}^{T}$, and seek a video
$\hat{V}=\{\hat{v}_t\}_{t=1}^{T}$ in which the subject of $x_{\mathrm{ref}}$ performs
$\mathcal{P}$ while retaining its appearance.

The base animator $F_{\theta}$ is a latent video diffusion model. An encoder maps
each frame to a latent $z$. A forward process adds Gaussian noise over diffusion
steps $\tau$,
\begin{equation}
  q(z_\tau \mid z_{\tau-1}) =
  \mathcal{N}\!\big(z_\tau;\,\sqrt{1-\beta_\tau}\,z_{\tau-1},\,\beta_\tau \mathbf{I}\big),
\end{equation}
and $F_{\theta}$ is trained to reverse it under conditioning
$c=(x_{\mathrm{ref}},\mathcal{P})$,
\begin{equation}
  \mathcal{L}(\theta)=
  \mathbb{E}_{z,\,c,\,\epsilon,\,\tau}
  \big\lVert\,\epsilon-\epsilon_{\theta}(z_\tau,\tau,c)\,\big\rVert_2^2,
  \qquad \epsilon\sim\mathcal{N}(0,\mathbf{I}).
  \label{eq:ldm}
\end{equation}
At inference $F_{\theta}$ maps $(x_{\mathrm{ref}},\mathcal{P})$ to a video but does
not reliably preserve the reference identity as the pose departs from the reference.

Personalization adapts the base through a small per-subject low-rank update
$\Delta\theta$, yielding the adapted model $F_{\theta+\Delta\theta}$. Since neither
subject video nor ground-truth frames are available, we draw the supervision from a
pose-conditioned generator $G$, whose selected pseudo frames
$\mathcal{D}_{\mathrm{pseudo}}$ stand in for the absent subject data
(\cref{sec:method:pseudo}).

\subsection{Reference-only reward-guided generation}
\label{sec:method:pseudo}

\noindent\textbf{Pose-conditioned generator $G$.}
We build $G$ from FLUX.1 Kontext~\cite{flux2025kontext}, a FLUX.1~\cite{flux2024}
variant with strong reference and prior preservation, augmented with a pose
ControlNet~\cite{zhang2023controlnet}. We domain-tune $G$ for this task. Given the reference image
$x_{\mathrm{ref}}$ and a target pose, $G$ renders the subject in that pose while
preserving identity. During training, we freeze the base weights of $G$ and optimize
a low-rank LoRA injected into its DiT attention blocks. Each training step draws a
two-frame pair from an in-domain video: one frame serves as the reference, and $G$
must reconstruct the second frame from its 2D pose. The optimizer, schedule, and
trained layers are in Appendix~\ref{app:impl}.

\begin{figure}[t]
  \centering
  \includegraphics[width=\linewidth]{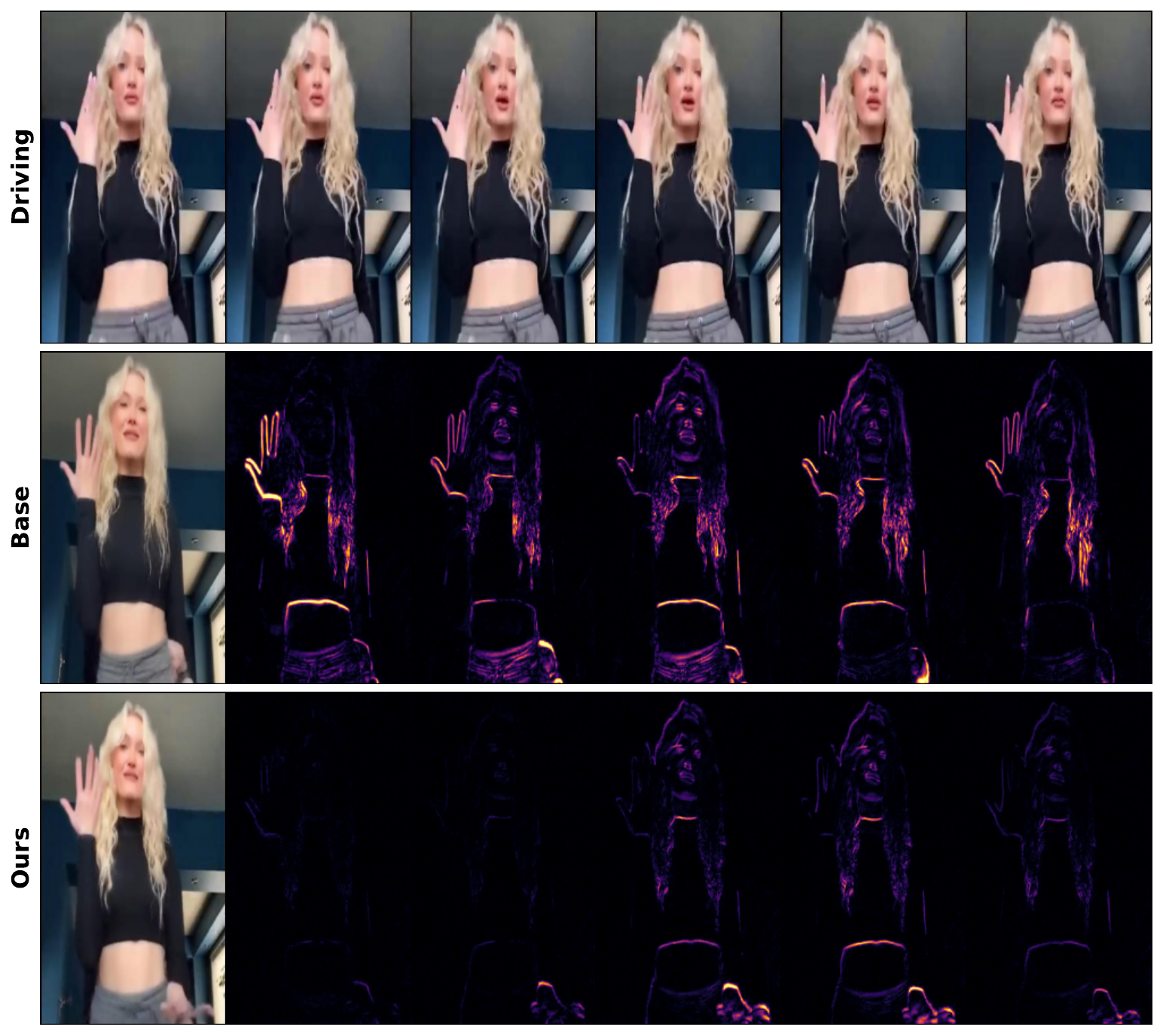}
  \caption{\textbf{Temporal stability.} \emph{Top:} the driving frames. For
  \emph{Base} (middle) and \emph{Ours} (bottom), the leftmost column is that
  method's output frame and the remaining columns are consecutive
  frame-difference maps over the \emph{same} frames. The non-personalized base
  produces widespread inter-frame residual across the face and torso (temporal
  flicker), whereas our adapted model confines residual to true motion boundaries
  and stays stable, reducing mean inter-frame residual by $60\%$ over the frames
  shown. Brighter indicates larger frame-to-frame change.}
  \label{fig:temporalstab}
\end{figure}

\noindent\textbf{Pose-targeted synthesis.}
From the driving sequence $\mathcal{P}$ we select $K$ key poses
$\mathcal{P}_K \subset \mathcal{P}$ by farthest-point sampling in keypoint space,
anchored at the reference pose.
This concentrates the supervision on the hard poses the video model $F_\theta$ must
render and spans the range of motion.
For each key pose $p_k \in \mathcal{P}_K$ the generator must produce one frame that
is faithful to the subject and of sufficient quality to supervise appearance.

\noindent\textbf{Reward-guided sampling.}
Rather than draw $N$ independent candidates and keep the best (best-of-$N$), we
sample the $N$ candidates as an interacting particle system and steer them while
they denoise.
We adapt inference-time particle steering~\cite{singhal2025fksteering,
ma2025inferencescaling} to our task with two novel changes specific to the
single-reference setting.
First, every reward is derived from the reference image alone. No ground-truth
target frame exists, so the guidance signal is self-supervised.
Second, we split the reward by role.

An in-loop steering reward
$R_{\mathrm{s}}(x)=w_{b}\,\mathrm{DINO}(x,x_{\mathrm{ref}})+w_{\sigma}\,\mathrm{sharp}(x)$
combines body and appearance similarity to the reference, read from a DINOv2
encoder~\cite{oquab2024dinov2}, with an image-sharpness term. At scheduled denoising
steps, the particles are resampled in proportion to $\exp(\lambda R_{\mathrm{s}})$,
so sampling compute concentrates on the trajectories that already resemble the
subject (\cref{alg:guided}).
We resample only in the back half of denoising. Early, high-noise particles lack the
structure for the reward to be informative, so the first half runs as a plain
deterministic ODE, and steering it would only collapse diversity. After each
resampling, a churn step re-noises the particles at strength $\eta$, so that
duplicated high-reward particles diverge rather than remaining identical under the
deterministic dynamics.

A final selection reward
$R_{\mathrm{f}}(x)=\mathrm{ArcFace}(x,x_{\mathrm{ref}})$~\cite{deng2019arcface}
then picks the surviving candidate with the highest face identity.
We steer in-loop with the dense, inexpensive DINO signal and reserve the sharper
but detector-sensitive ArcFace score for the single final choice, where it is most
reliable.
At an equal sampling budget, the reward-guided sampler improves identity and video
quality over best-of-$N$ (the no-steering rows of \cref{tab:rewardablation}).

The selected pair $(x_k^{*},p_k)$ enters the pseudo-supervision set
$\mathcal{D}_{\mathrm{pseudo}}=\{(x_k^{*},p_k)\}_{k=1}^{K}$.
Each selected frame captures a single pose. Replicated into a static clip for
adaptation (\cref{sec:method:tta}), it carries no inter-frame motion, unlike real
video frames captured from a subject in motion.
Reward-guided sampling and DINO- or identity-based rewards are each
established~\cite{singhal2025fksteering,wei2024personalizedrl,shen2026ipro}. Unlike
reward fine-tuning against ground-truth video~\cite{shen2026ipro,xue2025dancegrpo},
ours is training-free, uses only the reference image as its target, and acts at the
pseudo-supervision stage.
The novelty in our setting is that these frames are pose-targeted to the test
subject's driving sequence, then reward-steered against the reference before they
supervise.

\begin{algorithm}[t]
  \caption{\textsc{GuidedSample}: reward-guided pseudo frame generation ( one target pose) }
  \label{alg:guided}
  \small
  \begin{algorithmic}[1]
    \Statex \textbf{Input:} reference $x_{\mathrm{ref}}$, target pose $p_k$
    \Statex \textbf{Output:} pseudo frame $x_k^{*}$
    \Statex \textbf{Frozen model:} pose-conditioned generator $G$
    \Statex \textbf{Rewards:}
    $R_{\mathrm{s}}(x)=w_b\,\mathrm{DINO}(x,x_{\mathrm{ref}})+w_\sigma\,\mathrm{sharp}(x)$,\;
    $R_{\mathrm{f}}(x)=\mathrm{ArcFace}(x,x_{\mathrm{ref}})$
    \Statex \textbf{Hyperparameters:} particles $N$, denoising steps $S$, resample
    steps $\mathcal{R}\subseteq\{\lceil t_{\mathrm{start}}S\rceil,\dots,S\}$ (back
    half), temperature $\lambda$, churn $\eta$
    \State $\{z^{(i)}\}_{i=1}^{N} \sim \mathcal{N}(0,\mathbf{I})$
    \Comment{initialize $N$ particles}
    \For{$s = 1,\dots,S$}
      \State $z^{(i)} \gets \textsc{EulerStep}(G, z^{(i)}, x_{\mathrm{ref}}, p_k)\ \ \forall i$
      \Comment{deterministic ODE step}
      \If{$s \in \mathcal{R}$}
        \State $w_i \propto \exp\!\big(\lambda\, R_{\mathrm{s}}(\hat{x}_0^{(i)})\big)$
        \Comment{reward weights; $\hat{x}_0^{(i)}$ decoded from $z^{(i)}$}
        \State $\{z^{(i)}\} \sim \mathrm{Categorical}(w)$
        \Comment{resample $N$ particles with replacement, $\Pr(i)\!\propto\! w_i$}
        \State $z^{(i)} \gets \textsc{Renoise}(z^{(i)};\,\eta)\ \ \forall i$
        \Comment{churn: re-noise so duplicate particles diverge}
      \EndIf
    \EndFor
    \State \Return $x_k^{*} = \arg\max_i R_{\mathrm{f}}(x^{(i)}),\ \ x^{(i)}=\mathrm{decode}(z^{(i)})$
    \Comment{ArcFace pick}
  \end{algorithmic}
\end{algorithm}

\subsection{Test-time adaptation}
\label{sec:method:tta}

\noindent\textbf{Static-clip adaptation.}
We adapt only a low-rank update $\Delta\theta$ (LoRA \cite{hu2022lora}) of rank $r$
on $F_{\theta}$, and keep $\theta$ frozen.
We train $\Delta\theta$ by minimizing an adaptation loss
$\mathcal{L}_{\mathrm{adapt}}$ on $\mathcal{D}_{\mathrm{pseudo}}$.
Each adaptation step uses one selected frame $x_k^{*}$.
We replicate $x_k^{*}$ into a clip and hold the pose $p_k$ constant across every
frame of that clip, then train $F_{\theta+\Delta\theta}$ to reconstruct it.
The static clip carries no motion. Writing $\mathrm{clip}(x)=[x,\dots,x]$ for the
$T$-frame replication and $z_\tau^{k}$ for the noised latent of
$\mathcal{E}(\mathrm{clip}(x_k^{*}))$,
\begin{equation}
  \mathcal{L}_{\mathrm{adapt}}(\Delta\theta)=\sum_{k=1}^{K}
  \mathbb{E}_{\tau,\epsilon}\big\lVert\epsilon-
  \epsilon_{\theta+\Delta\theta}\big(z_\tau^{k},\tau,(x_{\mathrm{ref}},p_k)\big)\big\rVert_2^2,
  \label{eq:adapt}
\end{equation}
so the pose $p_k$ stays fixed across the clip and the temporal pathway sees no
motion during adaptation.

A multi-frame clip of distinct pseudo poses would re-introduce motion into the
temporal pathways of the unified self-attention DiT. That motion reproduces the
canonical motion-prior damage seen in video personalization
\cite{chefer2024stillmoving,zhao2024motiondirector,li2025personalvideo,li2025magicid}.
The static clip removes that signal, so adaptation updates appearance only.
Prior fixes that decouple spatial parameters from temporal parameters
\cite{ma2025followyourmotion} do not transfer to unified-attention DiTs, where one
attention operation mixes both.
We therefore control motion through the data rather than the parameterization.
Removing the static clip drops identity below the base model and worsens temporal
consistency, as we report in \cref{tab:buildup}. At inference, the adapted model
$F_{\theta+\Delta\theta}$ renders $x_{\mathrm{ref}}$ along the full driving sequence
$\mathcal{P}$ to produce $\hat{V}$.

\begin{algorithm}[b!]
  \caption{Test-time self-generated personalization}
  \label{alg:testtime}
  \small
  \begin{algorithmic}[1]
    \Statex \textbf{Input:} reference $x_{\mathrm{ref}}$, driving poses
    $\mathcal{P}=\{p_t\}_{t=1}^{T}$
    \Statex \textbf{Output:} personalized video $\hat{V}$
    \Statex \textbf{Frozen models:} base animator $F_{\theta}$, generator $G$
    \Statex \textbf{Trainable:} per-subject LoRA $\Delta\theta$ (rank $r$)
    \Statex \textbf{Hyperparameters:} key poses $K$
    \State $\mathcal{P}_K \gets \mathrm{FPS}(\mathcal{P}, K)$
    \Comment{select $K$ key poses by farthest-point sampling}
    \State $\mathcal{D}_{\mathrm{pseudo}} \gets \emptyset$
    \Comment{empty pseudo-supervision set}
    \For{$p_k \in \mathcal{P}_K$}
      \State $x_k^{*} \gets \textsc{GuidedSample}(x_{\mathrm{ref}}, p_k;\, G)$
      \Comment{reward-guided sampling from $G$, Alg.~\ref{alg:guided}}
      \State $\mathcal{D}_{\mathrm{pseudo}} \gets \mathcal{D}_{\mathrm{pseudo}} \cup \{(x_k^{*}, p_k)\}$
    \EndFor
    \State $\Delta\theta \gets \arg\min\limits_{\Delta\theta}
    \sum\limits_{(x,p)\in\mathcal{D}_{\mathrm{pseudo}}}
    \mathcal{L}_{\mathrm{adapt}}\!\big(F_{\theta+\Delta\theta};\,
    \mathrm{clip}(x), p\big)$
    \Comment{static clip of every pseudo frame, $\theta$ frozen}
    \State \Return $\hat{V} = F_{\theta+\Delta\theta}(x_{\mathrm{ref}}, \mathcal{P})$
  \end{algorithmic}
\end{algorithm}

\section{Experiments}
\label{sec:experiments}

\subsection{Setup}
\label{sec:exp:setup}

\noindent\textbf{Datasets.}
For in-domain training and evaluation we use the TikTok~\cite{jafarian2021tiktok}
and UBC Fashion~\cite{zablotskaia2019dwnet} benchmarks with their standard train and
test splits. To probe cross-domain transfer, we additionally evaluate zero-shot on
Unseen100-Val, a set of 100 in-the-wild videos spanning diverse subjects, motion,
lighting, and backgrounds (\cref{tab:crossdomain}). Full dataset details are in
Appendix~\ref{app:data}.

\noindent\textbf{Evaluation.}
We follow the evaluation protocol of RealisDance-DiT~\cite{zhou2025realisdancedit}
for all comparisons, covering reconstruction, perceptual, identity, and
video-quality metrics. Metric definitions
are in Appendix~\ref{app:metrics}, and the full protocol in Appendix~\ref{app:data:proto}.

\begin{figure*}[t!]
  \centering
  \includegraphics[width=\textwidth,height=0.25\textheight,keepaspectratio]{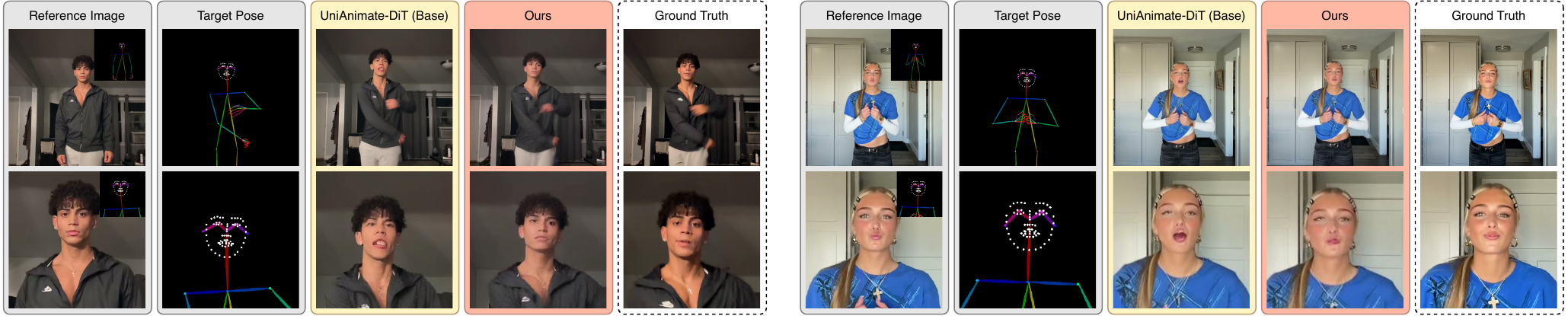}
  \caption{\textbf{Qualitative comparison} on hard poses. For two subjects we show
  the reference and driving pose alongside the base, our adapted model, and ground
  truth. Insets crop the face. The base reconstructs identity and garment detail
  inconsistently as the pose departs from the reference, while our adaptation
  preserves both.}
  \label{fig:qualitative}
\end{figure*}

\subsection{Implementation details}
\label{sec:exp:impl}

\noindent\textbf{Base animator.}
The base animator $F_{\theta}$ is UniAnimate-DiT~\cite{wang2025unianimatedit}, a
pose-driven human video DiT built on Wan2.1-I2V (14B)~\cite{wan2025}. It uses a
single unified 3D self-attention stack that models space and time jointly. Reference
conditioning enters through dual cross-attention with dedicated image-token
projections, and the target pose is injected as a residual at the latent entry. We
keep $\theta$ frozen and adapt only the low-rank update $\Delta\theta$ of rank $r$.

\noindent\textbf{Generator.}
We build the pose-conditioned generator $G$ on FLUX.1 Kontext~\cite{flux2025kontext}
with a pose ControlNet~\cite{zhang2023controlnet} and freeze its base. We train a
low-rank LoRA on its DiT attention blocks ourselves, on the in-domain training split
of each benchmark, with the two-frame reconstruction objective of
\cref{sec:method:pseudo}. We optimize with AdamW at a constant learning rate for
$4{,}000$ steps on a single H100 GPU. The LoRA rank, schedule, and trained
layers are in Appendix~\ref{app:impl}.

\noindent\textbf{Adaptation.}
By default we synthesize $K{=}5$ pseudo frames, each selected from $N{=}8$
reward-guided particles, and train the rank-$r$ LoRA on static clips of these
frames. The $K$ target poses, and the $K$ real frames of the Real-5 baseline,
are the same indices, chosen by farthest-point sampling over the driving poses and
seeded at the reference frame, so they span the pose range rather than the
timeline. We study how cost and quality scale with the number of fine-tuning steps in
Appendix~\ref{app:time}. Full hyperparameters are in Appendix~\ref{app:impl}.

\noindent\textbf{Test-time cost.}
\Cref{tab:testtimecost} reports the wall-clock and peak GPU memory of each stage on a
single H100. The method has two distinct costs, with different hardware requirements.
Generation and adaptation run once per reference image and take $9.3$ minutes
together, peaking at $62.1$~GiB. That peak is dominated by the frozen base model,
which accounts for $70\%$ of it, against $15\%$ for the low-rank update and its
optimizer state (\cref{tab:memorybreakdown}).
Inference is the recurring cost. It is unchanged by adaptation, since merging
the update adds no parameters at inference, and it peaks at $21.6$~GiB, within a
$24$~GB card. The two costs separate in deployment: the one-time stage runs
server-side and produces a
$1.1$~GiB adapter, after which inference proceeds at base cost.
Appendix~\ref{app:cost} gives the measurement protocol and the full breakdown.

\subsection{Comparison with the state of the art}
\label{sec:exp:sota}

\noindent\textbf{In-domain.}
\Cref{tab:main} compares against eight pose-driven animators on TikTok and Fashion dataset.
Our test-time adaptation improves the UniAnimate-DiT base on every metric and gives
the best result on most metrics. The largest gain is on video quality: FVD on TikTok
falls below every prior method.
Three aspects of this result are worth mentioning: the gain lands on the metric
that personalization usually degrades, it requires no subject video, and
reconstruction and identity improve in step rather than trading against video
quality. \Cref{fig:qualitative} shows
the effect on hard poses, where the base and competing animators reconstruct
identity inconsistently while our adapted model preserves the face and garment.
Additional qualitative results are in Appendix~\ref{app:qual}.

\begin{table}[tbp]
  \centering
  \caption{\textbf{Main comparison} on TikTok / Fashion (each cell TikTok/Fashion).
  StableAnimator is quoted from its paper, which reports neither FID nor FID-VID and does
  not evaluate on UBC Fashion.}
  \label{tab:main}
  \scriptsize
  \setlength{\tabcolsep}{2pt}
  \renewcommand{\arraystretch}{1.35}
  \resizebox{\linewidth}{!}{%
  \begin{tabular}{@{}lcccccc@{}}
    \toprule
    Model & PSNR$\uparrow$ & SSIM$\uparrow$ & LPIPS$\downarrow$ &
    FID$\downarrow$ & FVD$\downarrow$ & FID-VID$\downarrow$ \\
    \midrule
    MusePose~\cite{tmelyralab2024musepose} & 18.20/22.20 & 0.757/0.896 & 0.248/0.067 &
    41.99/14.95 & 532.75/96.17 & 14.60/10.94 \\
    Moore-AA~\cite{moorethreads2024mooreanimateanyone} & 18.62/20.83 & 0.764/0.880 & 0.230/0.093 &
    37.28/27.74 & 501.22/149.66 & 12.40/10.13 \\
    ControlNeXt~\cite{peng2024controlnext} & 16.31/18.48 & 0.728/0.853 & 0.296/0.132 &
    33.48/13.82 & 548.01/143.02 & 28.23/14.60 \\
    Animate-X~\cite{AnimateX2025} & 16.71/22.15 & 0.743/0.893 & 0.285/0.069 &
    32.77/10.11 & 508.63/70.47 & 17.47/11.87 \\
    MimicMotion~\cite{zhang2025mimicmotion} & 19.30/23.80 & 0.751/0.913 & 0.220/0.061 &
    34.88/15.40 & 472.51/80.89 & 9.30/8.17 \\
    RealisDance-DiT~\cite{zhou2025realisdancedit} & 17.55/23.33 & 0.717/0.908 & 0.261/\textbf{0.053} &
    30.39/10.81 & 458.81/72.94 & --/-- \\
    MTVCraft~\cite{ding2025mtvcraft} & 19.84/23.90 & 0.779/0.923 &
    0.217/0.057 & \textbf{20.70}/8.74 &
    276.65/64.88 & \textbf{7.31}/\textbf{4.41} \\
    StableAnimator~\cite{tu2025stableanimator} & 20.66/-- & 0.801/-- &
    0.232/-- & --/-- &
    140.62/-- & --/-- \\
    \midrule
    UniAnimate-DiT (Base)~\cite{wang2025unianimatedit} & 19.35/23.52 & 0.765/0.907 & 0.235/0.060 &
    28.47/12.79 & 402.14/88.36 & 9.12/6.12 \\
    \textbf{UA-DiT Ours} & \textbf{21.20/24.56} &
    \textbf{0.847/0.931} & \textbf{0.206}/0.059 &
    26.74/\textbf{8.71} &
    \textbf{133.5/53.88} & 8.42/5.79 \\
    \bottomrule
  \end{tabular}}
\end{table}

\begin{table}[tbp]
  \centering
  \caption{\textbf{Quantitative comparison} on cross-domain performance on the
  Unseen100-Val set after training on TikTok. Off the training distribution, our
  adaptation reaches the only FVD below $100$ ($92.5$) and the lowest dtSSD,
  while every other method stays above $125$.}
  \label{tab:crossdomain}
  \scriptsize
  \setlength{\tabcolsep}{2pt}
  \renewcommand{\arraystretch}{1.5}
  \resizebox{\linewidth}{!}{%
  \begin{tabular}{@{}l@{\hspace{3pt}}|ccccccc@{}}
    \toprule
    Model & PSNR$\uparrow$ & SSIM$\uparrow$ & LPIPS$\downarrow$ &
    FaceID$\uparrow$ & BodySIM$\uparrow$ & FVD$\downarrow$ & dtSSD$\downarrow$ \\
    \midrule
    MusePose~\cite{tmelyralab2024musepose} & 20.25 & 0.767 & 0.217 & 0.498 & 0.877 & 131.1 & 295.2 \\
    Moore-AA~\cite{moorethreads2024mooreanimateanyone} & 18.67 & 0.690 & 0.356 & 0.402 & 0.820 & 229.3 & 435.9 \\
    ControlNeXt~\cite{peng2024controlnext} & 17.85 & 0.720 & 0.328 & 0.534 & 0.854 & 226.1 & 1077.1 \\
    Animate-X~\cite{AnimateX2025} & 17.51 & 0.709 & 0.368 & 0.440 & 0.830 & 198.6 & 371.0 \\
    MimicMotion~\cite{zhang2025mimicmotion} & 17.28 & 0.704 & 0.346 & 0.488 & 0.874 & 219.6 & 769.9 \\
    RealisDance-DiT~\cite{zhou2025realisdancedit} & 18.43 & 0.663 & 0.339 & 0.574 & 0.885 & 181.5 & 594.4 \\
    \midrule
    UniAnimate-DiT (Base)~\cite{wang2025unianimatedit} & 21.19 & 0.796 & 0.206 & 0.632 & 0.903 & 125.5 & 207.9 \\
    \textbf{UniAnimate-DiT: Ours} & \textbf{21.45} & \textbf{0.804} &
    \textbf{0.203} & \textbf{0.663} & \textbf{0.904} & \textbf{92.5} &
    \textbf{190.0} \\
    \bottomrule
  \end{tabular}}
\end{table}

\noindent\textbf{Out-of-domain.}
Trained on TikTok and evaluated on the out-of-domain Unseen100-Val set
(\cref{tab:crossdomain}), the adaptation transfers. It leads all methods and
improves over the base on every metric, cutting FVD by roughly a quarter, with the
prior animators further behind. The advantage holds under domain shift
because the supervision is generated for the test subject rather than learned from
the training distribution, so it does not inherit a dataset-specific prior.

\noindent\textbf{Cross-reenactment.}
When one subject is driven by another subject's motion, no aligned ground truth
exists, so we measure identity against the reference (\cref{tab:crossreenact}).
Adaptation improves appearance fidelity (FaceID, BodySIM) while pose obedience
(PoseAcc) is essentially unchanged, since the pose-conditioning pathway stays
frozen. The adaptation thus binds appearance without disturbing the motion pathway,
the asymmetry the static-clip recipe is designed to produce.

\begin{table}[tbp]
  \centering
  \caption{\textbf{Cross-reenactment} (Unseen100-Val). Each subject is driven by a
  different subject's motion, so no aligned ground truth exists; we measure identity
  against the reference (FaceID, BodySIM) and pose obedience against the driving
  sequence (PoseAcc). Adaptation raises appearance fidelity while leaving pose
  obedience essentially unchanged, as the pose-conditioning pathway is frozen.}
  \label{tab:crossreenact}
  \scriptsize
  \setlength{\tabcolsep}{6pt}
    \begin{adjustbox}{max width=\columnwidth}
\begin{tabular}{@{}lccc@{}}
    \toprule
    Method & FaceID$\uparrow$ & BodySIM$\uparrow$ & PoseAcc$\uparrow$ \\
    \midrule
    UniAnimate-DiT (Base) & 0.431 & 0.745 & 0.817 \\
    \textbf{Ours} & \textbf{0.571} & \textbf{0.791} & \textbf{0.822} \\
    \midrule
    $\Delta$ & $+0.140$ & $+0.046$ & $+0.005$ \\
    \bottomrule
  \end{tabular}
  \end{adjustbox}
\end{table}

\begin{figure*}[tp]
  \centering
  \begin{subfigure}[b]{0.50\linewidth}
    \centering
    \resizebox{\linewidth}{!}{%
\begin{tikzpicture}
  \begin{axis}[
    width=\linewidth, height=4.0cm,
    xlabel={$K$}, xlabel near ticks,
    xtick={1,3,5,9,21}, xmin=0, xmax=22,
    axis y line*=left, axis x line=bottom,
    ylabel={FaceID $\uparrow$}, ymin=0.58, ymax=0.64,
    ylabel style={text=teal}, yticklabel style={text=teal},
    ylabel near ticks,
  ]
  \addplot[teal,mark=*,thick] coordinates
    {(1,0.590)(3,0.605)(5,0.617)(9,0.600)(21,0.626)};
  \end{axis}
  \begin{axis}[
    width=\linewidth, height=4.0cm,
    axis y line*=right, axis x line=none,
    ylabel={FVD $\downarrow$}, ymin=120, ymax=190,
    xmin=0, xmax=22,
    ylabel style={text=orange}, yticklabel style={text=orange},
    ylabel near ticks,
  ]
  \addplot[orange,mark=triangle*,dashed,thick] coordinates
    {(1,148.3)(3,178.6)(5,133.5)(9,152.6)(21,134.0)};
  \end{axis}
\end{tikzpicture}%
}
    \caption{Pseudo-frame count $K$}
    \label{fig:ksweep}
  \end{subfigure}
  \hfill
  \begin{subfigure}[b]{0.46\linewidth}
    \centering
    \begin{tikzpicture}
    \def\R{1.9}
    \foreach \ang/\lab in {90/PSNR,38.571/SSIM,-12.857/LPIPS,-64.286/FaceID,-115.714/BodySIM,-167.143/FVD,-218.571/dtSSD}{
      \draw[gray!40] (0,0) -- (\ang:\R);
      \node[font=\scriptsize] at (\ang:{\R+0.42}) {\lab};
    }
    \foreach \r in {0.25,0.5,0.75,1.0}{
      \draw[gray!20] (90:{\r*\R})
        \foreach \ang in {38.571,-12.857,-64.286,-115.714,-167.143,-218.571}
          {-- (\ang:{\r*\R})} -- cycle;
    }
    \draw[teal,fill=teal!12,fill opacity=0.35,very thick]
      (90:{1.0*\R}) -- (38.571:{0.947*\R}) -- (-12.857:{0.97*\R})
      -- (-64.286:{1.0*\R}) -- (-115.714:{0.953*\R}) -- (-167.143:{1.0*\R})
      -- (-218.571:{1.0*\R}) -- cycle;
    \draw[purple,fill=purple!10,fill opacity=0.30,thick,dotted]
      (90:{0.1*\R}) -- (38.571:{0.1*\R}) -- (-12.857:{0.1*\R})
      -- (-64.286:{0.46*\R}) -- (-115.714:{0.1*\R}) -- (-167.143:{0.1*\R})
      -- (-218.571:{0.1*\R}) -- cycle;
    \draw[orange,thick,dashed]
      (90:{0.703*\R}) -- (38.571:{0.947*\R}) -- (-12.857:{0.91*\R})
      -- (-64.286:{0.91*\R}) -- (-115.714:{1.0*\R}) -- (-167.143:{0.815*\R})
      -- (-218.571:{0.781*\R}) -- cycle;
    \draw[gray,thick]
      (90:{0.481*\R}) -- (38.571:{0.259*\R}) -- (-12.857:{0.16*\R})
      -- (-64.286:{0.1*\R}) -- (-115.714:{0.811*\R}) -- (-167.143:{0.437*\R})
      -- (-218.571:{0.369*\R}) -- cycle;
    \draw[teal,very thick] (-2.65,-2.45) -- (-2.35,-2.45);
    \node[right,font=\scriptsize] at (-2.35,-2.45) {Ours};
    \draw[orange,thick,dashed] (-1.55,-2.45) -- (-1.25,-2.45);
    \node[right,font=\scriptsize] at (-1.25,-2.45) {Real-5};
    \draw[purple,thick,dotted] (-0.25,-2.45) -- (0.05,-2.45);
    \node[right,font=\scriptsize] at (0.05,-2.45) {Self-distill};
    \draw[gray,thick] (1.55,-2.45) -- (1.85,-2.45);
    \node[right,font=\scriptsize] at (1.85,-2.45) {Base};
  \end{tikzpicture}
    \caption{Supervision source}
    \label{fig:radar}
  \end{subfigure}
  \caption{\textbf{Pseudo-frame count and supervision source.} (a) FVD (orange) is
  lowest at $K{=}5$; FaceID (teal) dips at $K{=}9$ and peaks at $K{=}21$ at a similar
  FVD (values from \cref{tab:kablation}). (b) Seven-metric comparison of supervision sources (outward $=$
  better, per-axis min--max over the rows of \cref{tab:source}): Ours forms nearly the outer envelope, while self-distillation collapses on body similarity and temporal
  consistency.}
  \label{fig:analysis}
\end{figure*}
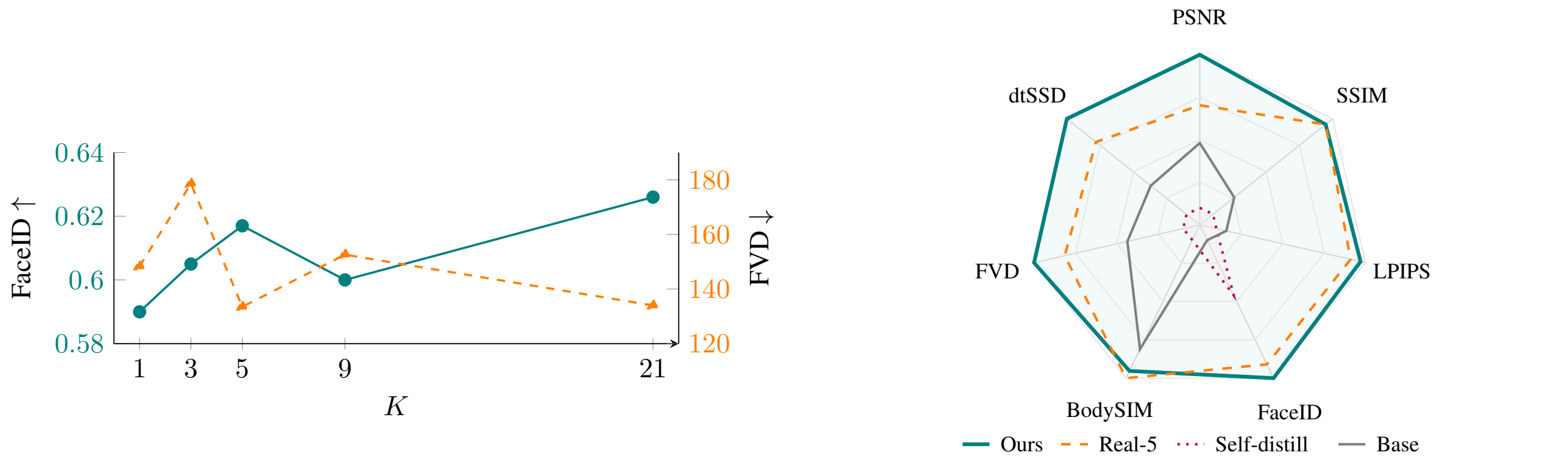

\subsection{Ablations and analysis}
\label{sec:exp:analysis}

\noindent\textbf{Contribution of each component.}
\Cref{tab:buildup} removes one component at a time. Removing the reward-guided frame
selection raises FVD as lower-quality pseudo frames enter adaptation. Dropping the static clip
is the most damaging: identity falls and temporal consistency degrades past the base
itself (dtSSD $369.0\rightarrow416.3$). This confirms that multi-pose clips push
motion back into the temporal pathway.

The full system also improves on adaptation
on five \emph{real} subject frames (Real-5) in FVD while using none, and matches
it on identity, inverting the usual trade-off. The standalone generator $G$ attains the
highest per-frame identity but has no temporal model and cannot produce coherent
video.

\noindent\textbf{Supervision source.}
Holding the adaptation fixed and varying only the supervision source isolates what
drives the gain (\cref{tab:source}). Self-distillation adapts the base on its own outputs. It
does not improve reconstruction and worsens temporal consistency, pushing dtSSD above the
base, and scarcely improves identity, since those outputs carry the motion
artifacts of a video model. Real frames help more, and our clean, static
pseudo-GT from $G$ helps most, cutting FVD to $133.5$.
\Cref{fig:radar} summarizes this
across all seven metrics.

\begin{table}[tbp]
  \centering
  \caption{\textbf{Component ablation.} Each designed component is removed and
  compared against real-frame supervision (Real-5) and the unadapted base. FaceID
  and BodySIM are over the full video. Generator $G$ is a per-frame reference
  with no temporal model, so it lacks FVD/dtSSD and is not ranked.}
  \label{tab:buildup}
  \scriptsize
  \setlength{\tabcolsep}{2pt}
  \renewcommand{\arraystretch}{1.5}
  \resizebox{\linewidth}{!}{%
  \begin{tabular}{@{}lccccccc@{}}
    \toprule
    Variant & PSNR$\uparrow$ & SSIM$\uparrow$ & LPIPS$\downarrow$ &
    FaceID$\uparrow$(long) & BodySIM$\uparrow$(long) & FVD$\downarrow$ &
    dtSSD$\downarrow$ \\
    \midrule
    Real-5 + recipe & 20.81 & \underline{0.847} & \underline{0.208} &
    \underline{0.615} & \textbf{0.888} & 157.3 &
    \underline{375.1} \\
    \midrule
    Ours (full) & \textbf{21.20} & \textbf{0.847} & \textbf{0.206} &
    \textbf{0.617} & \underline{0.887} & \textbf{133.5} & \textbf{369.0} \\
    w/o selection & \underline{20.92} & \underline{0.847} & \underline{0.208} &
    0.613 & 0.884 & 163.8 & 375.7 \\
    w/o static-clip & 20.80 & 0.842 & \underline{0.208} & 0.583 & 0.872 &
    \underline{151.9} & 416.3 \\
    Generator $G$ & 18.74 & 0.806 & 0.266 & 0.640 & 0.864 & $-$ & $-$ \\
    \midrule
    UniAnimate-DiT (Base) & 20.52 & 0.834 & 0.233 & 0.597 & 0.884 &
    206.0 & 386.6 \\
    \bottomrule
  \end{tabular}}
\end{table}

\begin{table}[tbp]
  \centering
  \caption{\textbf{Supervision source.} All rows share the adaptation
  recipe; only the \emph{supervision source} changes. FaceID and BodySIM are over
  the full video, on the TikTok dataset. The standalone quality of $G$ appears in
  \cref{tab:buildup}.}
  \label{tab:source}
  \scriptsize
  \setlength{\tabcolsep}{2pt}
  \renewcommand{\arraystretch}{1.5}
  \resizebox{\linewidth}{!}{%
  \begin{tabular}{@{}lccccccc@{}}
    \toprule
    Supervision source & PSNR$\uparrow$ & SSIM$\uparrow$ & LPIPS$\downarrow$ &
    FaceID$\uparrow$(long) & BodySIM$\uparrow$(long) & FVD$\downarrow$ &
    dtSSD$\downarrow$ \\
    \midrule
    UniAnimate-DiT (Base) & 20.52 & 0.834 & 0.233 & 0.597 & 0.884 & 206.0 & 386.6 \\
    Self-distillation (base's own frames) & 20.02 & 0.831 & 0.235 &
    0.605 & 0.869 & 249.4 & 394.1 \\
    Reference image only (no $G$) & 21.14 & \textbf{0.848} & \textbf{0.205} &
    0.608 & 0.885 & 139.0 & 380.5 \\
    Real-5 frames & 20.81 & 0.847 & 0.208 & 0.615 & \textbf{0.888} &
    157.3 & 375.1 \\
    \textbf{Ours ($G$ pseudo-GT)} & \textbf{21.20} & 0.847 &
    0.206 & \textbf{0.617} & 0.887 & \textbf{133.5} & \textbf{369.0} \\
    \bottomrule
  \end{tabular}}
\end{table}

\noindent\textbf{Number of pseudo frames.}
FVD is lowest at $K{=}5$, our default, and both FVD and identity degrade at $K{=}9$ as
extra frames expose generator artifacts (\cref{fig:ksweep}, \cref{tab:kablation}).
A larger $K{=}21$ reaches the highest full-video FaceID ($0.626$) but trades away the
balanced result, while a small $K$ under-covers the motion the model must render.
The fine-tuning-time and generator ablations are in Appendix~\ref{app:ablation}.

\begin{table}[tbp]
  \centering
  \caption{\textbf{Number of pseudo frames $K$.} We sweep the number of
  self-generated supervision frames used for per-subject adaptation. FaceID and
  BodySIM are over the full video; $K{=}5$ is the best balanced configuration.}
  \label{tab:kablation}
  \scriptsize
  \setlength{\tabcolsep}{2pt}
  \renewcommand{\arraystretch}{1.5}
  \resizebox{\linewidth}{!}{%
  \begin{tabular}{@{}lcccccccc@{}}
    \toprule
    $K$ & PSNR$\uparrow$ & SSIM$\uparrow$ & LPIPS$\downarrow$ &
    FaceID$\uparrow$(long) & BodySIM$\uparrow$(long) & FVD$\downarrow$ &
    FID$\downarrow$ & dtSSD$\downarrow$ \\
    \midrule
    1 & 20.33 & 0.836 & 0.232 & 0.590 & 0.875 & 148.3 & \textbf{25.40} & 388.5 \\
    3 & 20.93 & 0.845 & 0.210 & 0.605 & \underline{0.881} & 178.6 & 27.82 &
    \underline{377.5} \\
    5 (ours) & \textbf{21.20} & \textbf{0.847} & \textbf{0.206} &
    \underline{0.617} & \textbf{0.887} & \textbf{133.5} & \underline{26.74} &
    \textbf{369.0} \\
    9 & 20.79 & 0.842 & 0.212 & 0.600 & 0.872 & 152.6 & 28.46 & 396.0 \\
    21 & \underline{21.16} & \underline{0.846} & \underline{0.209} &
    \textbf{0.626} & \underline{0.881} & \underline{134.0} & 27.64 & 377.7 \\
    \bottomrule
  \end{tabular}}
\end{table}

\noindent\textbf{Steering and selection rewards.}
The reward-guided sampler reads two reference-only rewards in distinct roles, and
\cref{tab:rewardablation} shows both assignments matter. Its no-steering (i.i.d.)
rows are the best-of-$N$ baseline, which in-loop steering improves on at an equal
budget. The dense but coarse DINO reward steers the particle population best. The sharp but detector-sensitive ArcFace
reward is unreliable as an in-loop signal on the occluded, fast-moving faces typical
of dance, but is decisive as the final selection. Steering with DINO and selecting
with ArcFace is therefore our default.

\begin{table}[t]
  \centering
  \caption{\textbf{Reward-design ablation} (on a tuning subset). The two
  reference-only rewards assigned to in-loop steering versus final selection.
  Steering with DINO and selecting with ArcFace is best.}
  \label{tab:rewardablation}
  \scriptsize
  \setlength{\tabcolsep}{3pt}
  \renewcommand{\arraystretch}{1.4}
  \resizebox{\linewidth}{!}{%
  \begin{tabular}{@{}llcccc@{}}
    \toprule
    Steer (in-loop) & Select (final) & PSNR$\uparrow$ & FaceID$\uparrow$ &
    BodySIM$\uparrow$ & FVD$\downarrow$ \\
    \midrule
    none (i.i.d.) & DINO    & 21.48 & 0.685 & 0.916 & 713 \\
    none (i.i.d.) & ArcFace             & 22.42 & 0.695 & 0.918 & 538 \\
    DINO          & DINO                & 22.40 & 0.661 & 0.921 & 516 \\
    \textbf{DINO} & \textbf{ArcFace}    & \textbf{22.80} & \textbf{0.711} &
    \textbf{0.928} & 496 \\
    ArcFace & ArcFace & 21.07 & 0.655 & 0.922 & \textbf{429} \\
    \bottomrule
  \end{tabular}}
\end{table}

\begin{table}[t]
  \centering
  \caption{\textbf{Test-time cost per reference image} on a single H100 (TikTok,
  $81$-frame inference, adaptation at $832{\times}480$, generation at
  $576{\times}1024$). The one-time peak is the \emph{maximum} over the two stages, not
  their sum. Of it, $43.6$~GiB is the frozen base and $9.2$~GiB our update
  (\cref{tab:memorybreakdown}). Inference costs what the base costs, as the merged
  adapter adds no parameters. Protocol in App.~\ref{app:cost}.}
  \label{tab:testtimecost}
  \scriptsize
  \renewcommand{\arraystretch}{1.5}
  \setlength{\tabcolsep}{4pt}
  \begin{adjustbox}{max width=\columnwidth}
\begin{tabular}{@{}lcc@{}}
    \toprule
    Stage & Time & Peak GPU (GiB) \\
    \midrule
    \multicolumn{3}{@{}l}{\emph{Once per reference image}} \\
    \quad Pseudo-frame generation ($K{=}5$, $N{=}8$) & $9.2$\,min  & 41.7 \\
    \quad Per-subject LoRA adaptation ($5$ steps)    & $5.8$\,s  & 62.1 \\
    \quad\quad frozen base model                     & --   & 43.6 \\
    \quad\quad \textbf{our update} + optimizer state & --   & \textbf{9.2} \\
    \quad\quad activations, workspace, reserve       & --   & 9.4  \\
    \quad \textbf{Total}                             & \textbf{$9.3$\,min} & \textbf{62.1} \\
    \addlinespace[2pt]
    \multicolumn{3}{@{}l}{\emph{Per video}} \\
    \quad Inference (ours and base)                  & $1.8$\,min  & 21.6 \\
    \bottomrule
  \end{tabular}
  \end{adjustbox}
\end{table}

\noindent\textbf{Temporal stability.}
The gains do not come at the cost of motion. \Cref{fig:temporalstab} shows that
adaptation confines inter-frame residual to true motion boundaries rather than
spreading it across the face and torso. Identity improves without the flicker
that personalization usually introduces.

\section{Conclusion}
\label{sec:conclusion}

We personalize a pose-driven human video model to a subject from a single
reference image, with no subject video and no ground truth. A pose-conditioned
generator synthesizes pseudo-ground-truth frames at the poses the driving sequence
requires, steered and selected by reference-only rewards, and a per-subject LoRA adapts
the base animator on static clips of them. The static clip, which we adopt from prior work, supplies
subject appearance while carrying no motion, so the adaptation binds identity without
disturbing the temporal prior. Personalization then improves identity and
reconstruction and even improves video quality (FVD), the metric it usually degrades, and the gains transfer to out-of-domain subjects because the supervision is generated for the test subject rather than learned from a training distribution. The adapted model
surpasses its non-personalized base on every metric and matches or surpasses adaptation
on five real subject frames despite using none, inverting the trade-off personalization usually incurs.

\vspace{3pt}
\noindent\textbf{Limitations and future work.}
The method inherits the quality of its generator $G$, since a degraded pseudo frame
weakens the adaptation, so rejecting blurred or occluded frames is a simple
safeguard (App.~\ref{app:limit}). The reference-only rewards are fixed and
hand-weighted, and learning them would raise the ceiling $G$ imposes. The adapted model's own outputs could re-supervise $G$ in a closed loop. Because the
adaptation never touches the backbone, it should carry over to stronger animators as they appear.

{
    \small\ifdefined\nolinenumbers\nolinenumbers\fi
    \vskip-22pt
    \bibliographystyle{ieeenat_fullname}
    \bibliography{main}

\begin{thebibliography}{63}
\providecommand{\natexlab}[1]{#1}
\providecommand{\url}[1]{\texttt{#1}}
\expandafter\ifx\csname urlstyle\endcsname\relax
  \providecommand{\doi}[1]{doi: #1}\else
  \providecommand{\doi}{doi: \begingroup \urlstyle{rm}\Url}\fi

\bibitem[Ayupov et~al.(2025)Ayupov, Nakhodnov, Yaschenko, Kuznetsov, and
  Alanov]{ayupov2025dreamboothdpo}
Shamil Ayupov, Maksim Nakhodnov, Anastasia Yaschenko, Andrey Kuznetsov, and
  Aibek Alanov.
\newblock Dreamboothdpo: Improving personalized generation using direct
  preference optimization, 2025.

\bibitem[Balaji et~al.(2019)Balaji, Min, Bai, Chellappa, and
  Graf]{balaji2019conditional}
Yogesh Balaji, Martin~Renqiang Min, Bing Bai, Rama Chellappa, and Hans~Peter
  Graf.
\newblock Conditional gan with discriminative filter generation for
  text-to-video synthesis.
\newblock In \emph{IJCAI}, 2019.

\bibitem[Carreira and Zisserman(2017)]{Carreira2017QuoVA}
Jo{\~a}o Carreira and Andrew Zisserman.
\newblock Quo vadis, action recognition? a new model and the kinetics dataset.
\newblock In \emph{Proceedings of the IEEE Conference on Computer Vision and
  Pattern Recognition (CVPR)}, pages 4724--4733, 2017.

\bibitem[Chefer et~al.(2024)Chefer, Zada, Paiss, Ephrat, Tov, Rubinstein, Wolf,
  Dekel, Michaeli, and Mosseri]{chefer2024stillmoving}
Hila Chefer, Shiran Zada, Roni Paiss, Ariel Ephrat, Omer Tov, Michael
  Rubinstein, Lior Wolf, Tali Dekel, Tomer Michaeli, and Inbar Mosseri.
\newblock Still-moving: Customized video generation without customized video
  data.
\newblock \emph{ACM Transactions on Graphics}, 43\penalty0 (6):\penalty0
  244:1--244:11, 2024.

\bibitem[Chen et~al.(2025)Chen, Khurana, and Ramanan]{neurips2025dynamicnvs}
Kaihua Chen, Tarasha Khurana, and Deva Ramanan.
\newblock Reconstruct, inpaint, test-time finetune: Dynamic novel-view
  synthesis from monocular videos.
\newblock In \emph{Advances in Neural Information Processing Systems
  (NeurIPS)}, 2025.

\bibitem[Chen et~al.(2023)Chen, Hu, Li, Ruiz, Jia, Chang, and
  Cohen]{chen2023suti}
Wenhu Chen, Hexiang Hu, Yandong Li, Nataniel Ruiz, Xuhui Jia, Ming-Wei Chang,
  and William~W. Cohen.
\newblock Subject-driven text-to-image generation via apprenticeship learning.
\newblock In \emph{Advances in Neural Information Processing Systems
  (NeurIPS)}, pages 30286--30305, 2023.

\bibitem[Choi et~al.(2026)Choi, Lee, Nam, and Song]{choi2026priqa}
Inseong Choi, Siwoo Lee, Seung-Hun Nam, and Soohwan Song.
\newblock Pr-iqa: Partial-reference image quality assessment for
  diffusion-based novel view synthesis.
\newblock In \emph{Proceedings of the IEEE/CVF Conference on Computer Vision
  and Pattern Recognition (CVPR)}, pages 37301--37310, 2026.

\bibitem[Deng et~al.(2019)Deng, Guo, Xue, and Zafeiriou]{deng2019arcface}
Jiankang Deng, Jia Guo, Niannan Xue, and Stefanos Zafeiriou.
\newblock Arcface: Additive angular margin loss for deep face recognition.
\newblock In \emph{Proceedings of the IEEE/CVF Conference on Computer Vision
  and Pattern Recognition (CVPR)}, 2019.

\bibitem[Ding et~al.(2026)Ding, Hu, Guo, Zhang, Wang, He, Zhang, Wang, and
  Li]{ding2025mtvcraft}
Yanbo Ding, Xirui Hu, Zhizhi Guo, Yan Zhang, Xinrui Wang, Zhixiang He, Chi
  Zhang, Yali Wang, and Xuelong Li.
\newblock {MTVCraft}: Tokenizing 4d motion for arbitrary character animation.
\newblock In \emph{International Conference on Learning Representations
  (ICLR)}, 2026.

\bibitem[Gal et~al.(2023)Gal, Alaluf, Atzmon, Patashnik, Bermano, Chechik, and
  Cohen-Or]{gal2023textualinversion}
Rinon Gal, Yuval Alaluf, Yuval Atzmon, Or Patashnik, Amit~H. Bermano, Gal
  Chechik, and Daniel Cohen-Or.
\newblock An image is worth one word: Personalizing text-to-image generation
  using textual inversion.
\newblock In \emph{International Conference on Learning Representations
  (ICLR)}, 2023.

\bibitem[Han et~al.(2023)Han, Li, Zhang, Milanfar, Metaxas, and
  Yang]{han2023svdiff}
Ligong Han, Yinxiao Li, Han Zhang, Peyman Milanfar, Dimitris Metaxas, and Feng
  Yang.
\newblock {SVDiff}: Compact parameter space for diffusion fine-tuning.
\newblock In \emph{Proceedings of the IEEE/CVF International Conference on
  Computer Vision (ICCV)}, pages 7323--7334, 2023.

\bibitem[Heusel et~al.(2017)Heusel, Ramsauer, Unterthiner, Nessler, and
  Hochreiter]{heusel2017gans}
Martin Heusel, Hubert Ramsauer, Thomas Unterthiner, Bernhard Nessler, and Sepp
  Hochreiter.
\newblock Gans trained by a two time-scale update rule converge to a local nash
  equilibrium.
\newblock In \emph{NeurIPS}, 2017.

\bibitem[Hu et~al.(2022)Hu, Shen, Wallis, Allen-Zhu, Li, Wang, Wang, and
  Chen]{hu2022lora}
Edward~J. Hu, Yelong Shen, Phillip Wallis, Zeyuan Allen-Zhu, Yuanzhi Li, Shean
  Wang, Lu Wang, and Weizhu Chen.
\newblock {LoRA}: Low-rank adaptation of large language models.
\newblock In \emph{International Conference on Learning Representations
  (ICLR)}, 2022.

\bibitem[Hu(2024)]{hu2023animateanyone}
Li Hu.
\newblock Animate anyone: Consistent and controllable image-to-video synthesis
  for character animation.
\newblock In \emph{Proceedings of the IEEE/CVF Conference on Computer Vision
  and Pattern Recognition (CVPR)}, pages 8153--8163, 2024.

\bibitem[Huang et~al.(2026)Huang, Zhang, Hu, Wang, Zhang, Lu, Cheng, and
  Zhang]{huang2026wandancer}
Mingyang Huang, Peng Zhang, Li Hu, Guangyuan Wang, Ruoshi Zhang, Yi Lu, Gang
  Cheng, and Bang Zhang.
\newblock Wan-dancer: A hierarchical framework for minute-scale coherent
  music-to-dance generation.
\newblock \emph{arXiv preprint arXiv:2607.09581}, 2026.

\bibitem[Jafarian and Park(2021)]{jafarian2021tiktok}
Yasamin Jafarian and Hyun~Soo Park.
\newblock Learning high fidelity depths of dressed humans by watching social
  media dance videos.
\newblock In \emph{Proceedings of the IEEE/CVF Conference on Computer Vision
  and Pattern Recognition (CVPR)}, 2021.

\bibitem[Jiang et~al.(2024)Jiang, Wu, Yang, Si, Lin, Qiao, Loy, and
  Liu]{jiang2024videobooth}
Yuming Jiang, Tianxing Wu, Shuai Yang, Chenyang Si, Dahua Lin, Yu Qiao,
  Chen~Change Loy, and Ziwei Liu.
\newblock {VideoBooth}: Diffusion-based video generation with image prompts.
\newblock In \emph{Proceedings of the IEEE/CVF Conference on Computer Vision
  and Pattern Recognition (CVPR)}, pages 6689--6700, 2024.

\bibitem[Jin et~al.(2026)Jin, Won, Han, Dai, Luo, Baek, and
  Cho]{jin2026syntheticmotion}
Wonjoon Jin, Jiyun Won, Janghyeok Han, Qi Dai, Chong Luo, Seung-Hwan Baek, and
  Sunghyun Cho.
\newblock Learning to generate highly dynamic videos using synthetic motion
  data.
\newblock In \emph{Proceedings of the IEEE/CVF Conference on Computer Vision
  and Pattern Recognition (CVPR)}, pages 18331--18341, 2026.

\bibitem[Kumari et~al.(2023)Kumari, Zhang, Zhang, Shechtman, and
  Zhu]{kumari2023customdiffusion}
Nupur Kumari, Bingliang Zhang, Richard Zhang, Eli Shechtman, and Jun-Yan Zhu.
\newblock Multi-concept customization of text-to-image diffusion.
\newblock In \emph{Proceedings of the IEEE/CVF Conference on Computer Vision
  and Pattern Recognition (CVPR)}, pages 1931--1941, 2023.

\bibitem[Labs(2024)]{flux2024}
Black~Forest Labs.
\newblock Flux.
\newblock \url{https://github.com/black-forest-labs/flux}, 2024.

\bibitem[Labs et~al.(2025)Labs, Batifol, Blattmann, Boesel, Consul, Diagne,
  Dockhorn, English, English, Esser, Kulal, Lacey, Levi, Li, Lorenz,
  M{\"u}ller, Podell, Rombach, Saini, Sauer, and Smith]{flux2025kontext}
Black~Forest Labs, Stephen Batifol, Andreas Blattmann, Frederic Boesel, Saksham
  Consul, Cyril Diagne, Tim Dockhorn, Jack English, Zion English, Patrick
  Esser, Sumith Kulal, Kyle Lacey, Yam Levi, Cheng Li, Dominik Lorenz, Jonas
  M{\"u}ller, Dustin Podell, Robin Rombach, Harry Saini, Axel Sauer, and Luke
  Smith.
\newblock Flux.1 kontext: Flow matching for in-context image generation and
  editing in latent space, 2025.

\bibitem[Li et~al.(2025{\natexlab{a}})Li, Jiang, Xiao, Wang, Yi, Wu, and
  Cai]{li2025magicid}
Hengjia Li, Lifan Jiang, Xi Xiao, Tianyang Wang, Hongwei Yi, Boxi Wu, and Deng
  Cai.
\newblock {MagicID}: Hybrid preference optimization for id-consistent and
  dynamic-preserved video customization.
\newblock In \emph{Proceedings of the IEEE/CVF International Conference on
  Computer Vision (ICCV)}, pages 12737--12746, 2025{\natexlab{a}}.

\bibitem[Li et~al.(2025{\natexlab{b}})Li, Qiu, Zhang, Wang, Wei, Li, Zhang, Wu,
  and Cai]{li2025personalvideo}
Hengjia Li, Haonan Qiu, Shiwei Zhang, Xiang Wang, Yujie Wei, Zekun Li, Yingya
  Zhang, Boxi Wu, and Deng Cai.
\newblock {PersonalVideo}: High id-fidelity video customization without dynamic
  and semantic degradation.
\newblock In \emph{Proceedings of the IEEE/CVF International Conference on
  Computer Vision (ICCV)}, pages 19406--19416, 2025{\natexlab{b}}.

\bibitem[Li et~al.(2026)Li, Yu, Li, Yu, Adeli, Liu, and Wu]{li2026anylift}
Hongjie Li, Heng Yu, Jiaman Li, Hong-Xing Yu, Ehsan Adeli, C.~Karen Liu, and
  Jiajun Wu.
\newblock Anylift: Scaling motion reconstruction from internet videos via 2d
  diffusion.
\newblock In \emph{Proceedings of the IEEE/CVF Conference on Computer Vision
  and Pattern Recognition (CVPR)}, pages 13876--13886, 2026.

\bibitem[Lin et~al.(2025)Lin, Jiang, Yang, Zheng, Liang, Zhang, and
  Liu]{lin2025omnihuman}
Gaojie Lin, Jianwen Jiang, Jiaqi Yang, Zerong Zheng, Chao Liang, Yuan Zhang,
  and Jingtuo Liu.
\newblock {OmniHuman-1}: Rethinking the scaling-up of one-stage conditioned
  human animation models.
\newblock In \emph{Proceedings of the IEEE/CVF International Conference on
  Computer Vision (ICCV)}, pages 13847--13858, 2025.

\bibitem[Liu et~al.(2026)Liu, Sun, Ren, Broaddus, Huang, and
  Guigues]{liu2026had}
Xi Liu, Weiwei Sun, Zhou Ren, Chris Broaddus, Siyu Huang, and Laurent Guigues.
\newblock Had: Hallucination-aware diffusion priors for 3d reconstruction.
\newblock In \emph{Proceedings of the IEEE/CVF Conference on Computer Vision
  and Pattern Recognition (CVPR)}, pages 29781--29791, 2026.

\bibitem[Long et~al.(2026)Long, Wan, Nakhost, Lee, Pfister, and
  Ar{\i}k]{long2026vista}
Do~Xuan Long, Xingchen Wan, Hootan Nakhost, Chen-Yu Lee, Tomas Pfister, and
  Sercan~{\"O}. Ar{\i}k.
\newblock {VISTA}: A test-time self-improving video generation agent.
\newblock In \emph{Proceedings of the IEEE/CVF Conference on Computer Vision
  and Pattern Recognition (CVPR)}, pages 6021--6032, 2026.

\bibitem[Ma et~al.(2025{\natexlab{a}})Ma, Tong, Jia, Hu, Su, Zhang, Yang, Li,
  Jaakkola, Jia, and Xie]{ma2025inferencescaling}
Nanye Ma, Shangyuan Tong, Haolin Jia, Hexiang Hu, Yu-Chuan Su, Mingda Zhang,
  Xuan Yang, Yandong Li, Tommi Jaakkola, Xuhui Jia, and Saining Xie.
\newblock Scaling inference time compute for diffusion models.
\newblock In \emph{Proceedings of the Computer Vision and Pattern Recognition
  Conference (CVPR)}, pages 2523--2534, 2025{\natexlab{a}}.

\bibitem[Ma et~al.(2026)Ma, Liu, Zhu, Yang, Feng, Zhang, Yan, Li, Han, Qi, and
  Chen]{ma2025followyourmotion}
Yue Ma, Yulong Liu, Qiyuan Zhu, Xiangpeng Yang, Kunyu Feng, Xinhua Zhang,
  Zexuan Yan, Zhifeng Li, Sirui Han, Chenyang Qi, and Qifeng Chen.
\newblock {EffiVMT}: Video motion transfer via efficient spatial-temporal
  decoupled finetuning.
\newblock In \emph{International Conference on Learning Representations
  (ICLR)}, 2026.

\bibitem[Ma et~al.(2025{\natexlab{b}})Ma, Zhou, Wang, Yeh, Li, Yang, Dong,
  Keutzer, and Feng]{ma2024magicme}
Ze Ma, Daquan Zhou, Xue-She Wang, Chun-Hsiao Yeh, Xiuyu Li, Huanrui Yang, Zhen
  Dong, Kurt Keutzer, and Jiashi Feng.
\newblock Magic-me: Identity-specific video customized diffusion.
\newblock In \emph{Computer Vision -- ECCV 2024 Workshops}, pages 19--37.
  Springer, 2025{\natexlab{b}}.

\bibitem[{Moore Threads}(2024)]{moorethreads2024mooreanimateanyone}
{Moore Threads}.
\newblock Moore-animateanyone.
\newblock \url{https://github.com/MooreThreads/Moore-AnimateAnyone}, 2024.
\newblock Software repository.

\bibitem[Oquab et~al.(2024)Oquab, Darcet, Moutakanni, Vo, Szafraniec, Khalidov,
  Fernandez, Haziza, Massa, El-Nouby, Assran, Ballas, Galuba, Howes, Huang, Li,
  Misra, Rabbat, Sharma, Synnaeve, Xu, Jegou, Mairal, Labatut, Joulin, and
  Bojanowski]{oquab2024dinov2}
Maxime Oquab, Timoth{\'e}e Darcet, Th{\'e}o Moutakanni, Huy~V. Vo, Marc
  Szafraniec, Vasil Khalidov, Pierre Fernandez, Daniel Haziza, Francisco Massa,
  Alaaeldin El-Nouby, Mahmoud Assran, Nicolas Ballas, Wojciech Galuba, Russell
  Howes, Po-Yao Huang, Shang-Wen Li, Ishan Misra, Michael Rabbat, Vasu Sharma,
  Gabriel Synnaeve, Hu Xu, Herv{\'e} Jegou, Julien Mairal, Patrick Labatut,
  Armand Joulin, and Piotr Bojanowski.
\newblock {DINOv2}: Learning robust visual features without supervision.
\newblock \emph{Transactions on Machine Learning Research}, 2024.

\bibitem[Peng et~al.(2025)Peng, Wang, Zhang, Li, Yang, and
  Jia]{peng2024controlnext}
Bohao Peng, Jian Wang, Yuechen Zhang, Wenbo Li, Ming-Chang Yang, and Jiaya Jia.
\newblock Controlnext: Powerful and efficient control for image and video
  generation, 2025.

\bibitem[Ruiz et~al.(2023)Ruiz, Li, Jampani, Pritch, Rubinstein, and
  Aberman]{ruiz2023dreambooth}
Nataniel Ruiz, Yuanzhen Li, Varun Jampani, Yael Pritch, Michael Rubinstein, and
  Kfir Aberman.
\newblock Dreambooth: Fine tuning text-to-image diffusion models for
  subject-driven generation.
\newblock In \emph{Proceedings of the IEEE/CVF Conference on Computer Vision
  and Pattern Recognition (CVPR)}, pages 22500--22510, 2023.

\bibitem[Shen et~al.(2026)Shen, Jiang, Zhu, Li, Ge, Cao, and
  Zheng]{shen2026ipro}
Liao Shen, Wentao Jiang, Yiran Zhu, Jiahe Li, Tiezheng Ge, Zhiguo Cao, and Bo
  Zheng.
\newblock Identity-preserving image-to-video generation via reward-guided
  optimization.
\newblock In \emph{Proceedings of the IEEE/CVF Conference on Computer Vision
  and Pattern Recognition (CVPR)}, pages 27281--27290, 2026.

\bibitem[Singhal et~al.(2025)Singhal, Horvitz, Teehan, Ren, Yu, McKeown, and
  Ranganath]{singhal2025fksteering}
Raghav Singhal, Zachary Horvitz, Ryan Teehan, Mengye Ren, Zhou Yu, Kathleen
  McKeown, and Rajesh Ranganath.
\newblock A general framework for inference-time scaling and steering of
  diffusion models.
\newblock In \emph{Proceedings of the 42nd International Conference on Machine
  Learning (ICML)}, pages 55810--55827. PMLR, 2025.

\bibitem[Song et~al.(2026)Song, Yang, Zhao, Li, and Zhang]{song2026worldforge}
Chenxi Song, Yanming Yang, Tong Zhao, Ruibo Li, and Chi Zhang.
\newblock Taming video models for 3d and 4d generation via zero-shot camera
  control.
\newblock In \emph{Proceedings of the IEEE/CVF Conference on Computer Vision
  and Pattern Recognition (CVPR)}, pages 40352--40363, 2026.

\bibitem[Tan et~al.(2025)Tan, Gong, Wang, Zhang, Zheng, Zheng, Zheng, Chen, and
  Yang]{AnimateX2025}
Shuai Tan, Biao Gong, Xiang Wang, Shiwei Zhang, Dandan Zheng, Ruobing Zheng,
  Kecheng Zheng, Jingdong Chen, and Ming Yang.
\newblock Animate-x: Universal character image animation with enhanced motion
  representation.
\newblock In \emph{ICLR}, 2025.

\bibitem[Tong et~al.(2024)Tong, Li, Chen, Wu, and Zhou]{tmelyralab2024musepose}
Zhengyan Tong, Chao Li, Zhaokang Chen, Bin Wu, and Wenjiang Zhou.
\newblock Musepose: a pose-driven image-to-video framework for virtual human
  generation.
\newblock \emph{arXiv}, 2024.

\bibitem[Topalo{\u{g}}lu et~al.(2026)Topalo{\u{g}}lu, Li, Niemeyer, Navab,
  Tekalp, and Tombari]{topaloglu2026oraclegs}
Atakan Topalo{\u{g}}lu, Kunyi Li, Michael Niemeyer, Nassir Navab, A.~Murat
  Tekalp, and Federico Tombari.
\newblock Oraclegs: Grounding generative priors for sparse-view gaussian
  splatting.
\newblock In \emph{Proceedings of the IEEE/CVF Winter Conference on
  Applications of Computer Vision (WACV)}, pages 77--87, 2026.

\bibitem[Tu et~al.(2025)Tu, Xing, Han, Cheng, Dai, Luo, and
  Wu]{tu2025stableanimator}
Shuyuan Tu, Zhen Xing, Xintong Han, Zhi-Qi Cheng, Qi Dai, Chong Luo, and Zuxuan
  Wu.
\newblock Stableanimator: High-quality identity-preserving human image
  animation.
\newblock In \emph{CVPR}, 2025.

\bibitem[Unterthiner et~al.(2018)Unterthiner, van Steenkiste, Kurach, Marinier,
  Michalski, and Gelly]{unterthiner2019accurategenerativemodelsvideo}
Thomas Unterthiner, Sjoerd van Steenkiste, Karol Kurach, Raphael Marinier,
  Marcin Michalski, and Sylvain Gelly.
\newblock Towards accurate generative models of video: A new metric \&
  challenges.
\newblock \emph{arXiv preprint arXiv:1812.01717}, 2018.

\bibitem[Wan et~al.(2025)Wan, Wang, Ai, Wen, Mao, Xie, Chen, Yu, Zhao, Yang,
  Zeng, Wang, Zhang, Zhou, Wang, Chen, Zhu, Zhao, Yan, Huang, Feng, Zhang, Li,
  Wu, Chu, Feng, Zhang, Sun, Fang, Wang, Gui, Weng, Shen, Lin, Wang, Wang,
  Zhou, Wang, Shen, Yu, Shi, Huang, Xu, Kou, Lv, Li, Liu, Wang, Zhang, Huang,
  Li, Wu, Liu, Pan, Zheng, Hong, Shi, Feng, Jiang, Han, Wu, and Liu]{wan2025}
Team Wan, Ang Wang, Baole Ai, Bin Wen, Chaojie Mao, Chen-Wei Xie, Di Chen,
  Feiwu Yu, Haiming Zhao, Jianxiao Yang, Jianyuan Zeng, Jiayu Wang, Jingfeng
  Zhang, Jingren Zhou, Jinkai Wang, Jixuan Chen, Kai Zhu, Kang Zhao, Keyu Yan,
  Lianghua Huang, Mengyang Feng, Ningyi Zhang, Pandeng Li, Pingyu Wu, Ruihang
  Chu, Ruili Feng, Shiwei Zhang, Siyang Sun, Tao Fang, Tianxing Wang, Tianyi
  Gui, Tingyu Weng, Tong Shen, Wei Lin, Wei Wang, Wei Wang, Wenmeng Zhou, Wente
  Wang, Wenting Shen, Wenyuan Yu, Xianzhong Shi, Xiaoming Huang, Xin Xu, Yan
  Kou, Yangyu Lv, Yifei Li, Yijing Liu, Yiming Wang, Yingya Zhang, Yitong
  Huang, Yong Li, You Wu, Yu Liu, Yulin Pan, Yun Zheng, Yuntao Hong, Yupeng
  Shi, Yutong Feng, Zeyinzi Jiang, Zhen Han, Zhi-Fan Wu, and Ziyu Liu.
\newblock Wan: Open and advanced large-scale video generative models, 2025.

\bibitem[Wang et~al.(2025{\natexlab{a}})Wang, Zhang, Gao, Wang, Zhou, Zhang,
  Yan, and Sang]{wang2024unianimate}
Xiang Wang, Shiwei Zhang, Changxin Gao, Jiayu Wang, Xiaoqiang Zhou, Yingya
  Zhang, Luxin Yan, and Nong Sang.
\newblock {UniAnimate}: Taming unified video diffusion models for consistent
  human image animation.
\newblock \emph{Science China Information Sciences}, 68\penalty0 (10):\penalty0
  200103, 2025{\natexlab{a}}.

\bibitem[Wang et~al.(2025{\natexlab{b}})Wang, Zhang, Tang, Zhang, Gao, Wang,
  and Sang]{wang2025unianimatedit}
Xiang Wang, Shiwei Zhang, Longxiang Tang, Yingya Zhang, Changxin Gao, Yuehuan
  Wang, and Nong Sang.
\newblock Unianimate-dit: Human image animation with large-scale video
  diffusion transformer, 2025{\natexlab{b}}.

\bibitem[Wang et~al.(2004)Wang, Bovik, Sheikh, and Simoncelli]{SSIM}
Zhou Wang, Alan~C. Bovik, Hamid~R. Sheikh, and Eero~P. Simoncelli.
\newblock Image quality assessment: From error visibility to structural
  similarity.
\newblock \emph{IEEE Transactions on Image Processing}, 13\penalty0
  (4):\penalty0 600--612, 2004.

\bibitem[Wei et~al.(2024{\natexlab{a}})Wei, Zeng, Li, Yin, Duan, and
  Li]{wei2024personalizedrl}
Fanyue Wei, Wei Zeng, Zhenyang Li, Dawei Yin, Lixin Duan, and Wen Li.
\newblock Powerful and flexible: Personalized text-to-image generation via
  reinforcement learning.
\newblock In \emph{Computer Vision -- ECCV 2024}, pages 394--410. Springer,
  2024{\natexlab{a}}.

\bibitem[Wei et~al.(2024{\natexlab{b}})Wei, Zhang, Qing, Yuan, Liu, Liu, Zhang,
  Zhou, and Shan]{wei2024dreamvideo}
Yujie Wei, Shiwei Zhang, Zhiwu Qing, Hangjie Yuan, Zhiheng Liu, Yu Liu, Yingya
  Zhang, Jingren Zhou, and Hongming Shan.
\newblock {DreamVideo}: Composing your dream videos with customized subject and
  motion.
\newblock In \emph{Proceedings of the IEEE/CVF Conference on Computer Vision
  and Pattern Recognition (CVPR)}, pages 6537--6549, 2024{\natexlab{b}}.

\bibitem[Wu et~al.(2023)Wu, Ge, Wang, Lei, Gu, Shi, Hsu, Shan, Qie, and
  Shou]{wu2023tuneavideo}
Jay~Zhangjie Wu, Yixiao Ge, Xintao Wang, Stan~Weixian Lei, Yuchao Gu, Yufei
  Shi, Wynne Hsu, Ying Shan, Xiaohu Qie, and Mike~Zheng Shou.
\newblock {Tune-A-Video}: One-shot tuning of image diffusion models for
  text-to-video generation.
\newblock In \emph{Proceedings of the IEEE/CVF International Conference on
  Computer Vision (ICCV)}, pages 7623--7633, 2023.

\bibitem[Wu et~al.(2025)Wu, Zhang, Wang, Zhou, Zheng, Qi, Shan, and
  Li]{wu2025customcrafter}
Tao Wu, Yong Zhang, Xintao Wang, Xianpan Zhou, Guangcong Zheng, Zhongang Qi,
  Ying Shan, and Xi Li.
\newblock {CustomCrafter}: Customized video generation with preserving motion
  and concept composition abilities.
\newblock In \emph{Proceedings of the AAAI Conference on Artificial
  Intelligence}, pages 8469--8477, 2025.

\bibitem[Xu et~al.(2026)Xu, Zheng, Wang, Yu, Chen, Zhang, Zhou, Lee, Li, and
  Jiang]{xu2025hypermotion}
Shuolin Xu, Siming Zheng, Ziyi Wang, HC Yu, Jinwei Chen, Huaqi Zhang, Daquan
  Zhou, Tong-Yee Lee, Bo Li, and Peng-Tao Jiang.
\newblock Hypermotionx: The dataset and benchmark with dit-based pose-guided
  human image animation of complex motions, 2026.

\bibitem[Xu et~al.(2025)Xu, Wang, and Yu]{neurips2025fewglimpses}
Yan Xu, Yixing Wang, and Stella~X. Yu.
\newblock Novel view synthesis from a few glimpses via test-time natural video
  completion.
\newblock In \emph{Advances in Neural Information Processing Systems
  (NeurIPS)}, 2025.

\bibitem[Xu et~al.(2024)Xu, Zhang, Liew, Yan, Liu, Zhang, Feng, and
  Shou]{xu2024magicanimate}
Zhongcong Xu, Jianfeng Zhang, Jun~Hao Liew, Hanshu Yan, Jia-Wei Liu, Chenxu
  Zhang, Jiashi Feng, and Mike~Zheng Shou.
\newblock {MagicAnimate}: Temporally consistent human image animation using
  diffusion model.
\newblock In \emph{Proceedings of the IEEE/CVF Conference on Computer Vision
  and Pattern Recognition (CVPR)}, pages 1481--1490, 2024.

\bibitem[Xue et~al.(2025)Xue, Wu, Gao, Kong, Zhu, Chen, Liu, Liu, Guo, Huang,
  and Luo]{xue2025dancegrpo}
Zeyue Xue, Jie Wu, Yu Gao, Fangyuan Kong, Lingting Zhu, Mengzhao Chen, Zhiheng
  Liu, Wei Liu, Qiushan Guo, Weilin Huang, and Ping Luo.
\newblock Dancegrpo: Unleashing grpo on visual generation, 2025.

\bibitem[Yang et~al.(2023)Yang, Zeng, Yuan, and
  Li]{yang2023effectivewholebodyposeestimation}
Zhendong Yang, Ailing Zeng, Chun Yuan, and Yu Li.
\newblock Effective whole-body pose estimation with two-stages distillation.
\newblock In \emph{Proceedings of the IEEE/CVF International Conference on
  Computer Vision (ICCV) Workshops}, pages 4210--4220, 2023.

\bibitem[Zablotskaia et~al.(2019)Zablotskaia, Siarohin, Zhao, and
  Sigal]{zablotskaia2019dwnet}
Polina Zablotskaia, Aliaksandr Siarohin, Bo Zhao, and Leonid Sigal.
\newblock {DwNet}: Dense warp-based network for pose-guided human video
  generation.
\newblock In \emph{Proceedings of the British Machine Vision Conference
  (BMVC)}, 2019.

\bibitem[Zhang et~al.(2023)Zhang, Rao, and Agrawala]{zhang2023controlnet}
Lvmin Zhang, Anyi Rao, and Maneesh Agrawala.
\newblock Adding conditional control to text-to-image diffusion models.
\newblock In \emph{Proceedings of the IEEE/CVF International Conference on
  Computer Vision (ICCV)}, pages 3836--3847, 2023.

\bibitem[Zhang et~al.(2018)Zhang, Isola, Efros, Shechtman, and Wang]{LPIPS}
Richard Zhang, Phillip Isola, Alexei~A Efros, Eli Shechtman, and Oliver Wang.
\newblock The unreasonable effectiveness of deep features as a perceptual
  metric.
\newblock In \emph{CVPR}, pages 586--595, 2018.

\bibitem[Zhang et~al.(2025)Zhang, Gu, Wang, Wang, Cheng, Zhu, and
  Zou]{zhang2025mimicmotion}
Yuang Zhang, Jiaxi Gu, Li-Wen Wang, Han Wang, Junqi Cheng, Yuefeng Zhu, and
  Fangyuan Zou.
\newblock {MimicMotion}: High-quality human motion video generation with
  confidence-aware pose guidance.
\newblock In \emph{Proceedings of the 42nd International Conference on Machine
  Learning (ICML)}, pages 74896--74910. PMLR, 2025.

\bibitem[Zhao et~al.(2026)Zhao, Qi, Wang, Zheng, Lu, Chen, Xu, Wu, and
  Jiang]{zhao2025dynamictrl}
Haoyu Zhao, Zhongang Qi, Cong Wang, Qingping Zheng, Guansong Lu, Fei Chen, Hang
  Xu, Zuxuan Wu, and Yu-Gang Jiang.
\newblock {DynamiCtrl}: Rethinking the basic structure and the role of text for
  high-quality human image animation.
\newblock \emph{IEEE Transactions on Multimedia}, 2026.
\newblock Early Access.

\bibitem[Zhao et~al.(2024)Zhao, Gu, Wu, Zhang, Liu, Wu, Keppo, and
  Shou]{zhao2024motiondirector}
Rui Zhao, Yuchao Gu, Jay~Zhangjie Wu, David~Junhao Zhang, Jia-Wei Liu, Weijia
  Wu, Jussi Keppo, and Mike~Zheng Shou.
\newblock {MotionDirector}: Motion customization of text-to-video diffusion
  models.
\newblock In \emph{Computer Vision -- ECCV 2024}, pages 273--290. Springer,
  2024.

\bibitem[Zhou et~al.(2025)Zhou, Wu, Li, Wei, Fan, Chen, Jiang, and
  Wang]{zhou2025realisdancedit}
Jingkai Zhou, Yifan Wu, Shikai Li, Min Wei, Chao Fan, Weihua Chen, Wei Jiang,
  and Fan Wang.
\newblock Realisdance-dit: Simple yet strong baseline towards controllable
  character animation in the wild, 2025.

\bibitem[Zhu et~al.(2024)Zhu, Chen, Dai, Dong, Xu, Cao, Yao, Zhu, and
  Zhu]{zhu2024champ}
Shenhao Zhu, Junming~Leo Chen, Zuozhuo Dai, Zilong Dong, Yinghui Xu, Xun Cao,
  Yao Yao, Hao Zhu, and Siyu Zhu.
\newblock Champ: Controllable and consistent human image animation with {3D}
  parametric guidance.
\newblock In \emph{Computer Vision -- ECCV 2024}, pages 145--162. Springer,
  2024.

\end{thebibliography}
}

\clearpage
\appendix

\addtocontents{toc}{\protect\setcounter{tocdepth}{2}}
\onecolumn
\begin{center}
  {\large One Image Is Enough: Self-Generated Supervision for\\
   Personalizing Human Video Diffusion at Test Time}\\[0.8em]
  {\Large\bfseries Supplementary Material}\\[1.4em]
\end{center}

\begin{center}\textsc{\large Table of Contents}\end{center}
\noindent\rule{\linewidth}{0.4pt}\par
\setlength{\cftbeforesecskip}{13pt}
\setlength{\cftbeforesubsecskip}{5pt}
\vspace*{\fill}
\begingroup\hypersetup{linkcolor=black}
\makeatletter\@starttoc{toc}\makeatother
\endgroup
\vspace*{\fill}
\noindent\rule{\linewidth}{0.4pt}
\twocolumn

\begin{figure*}[t]
  \centering
  \includegraphics[width=\linewidth]{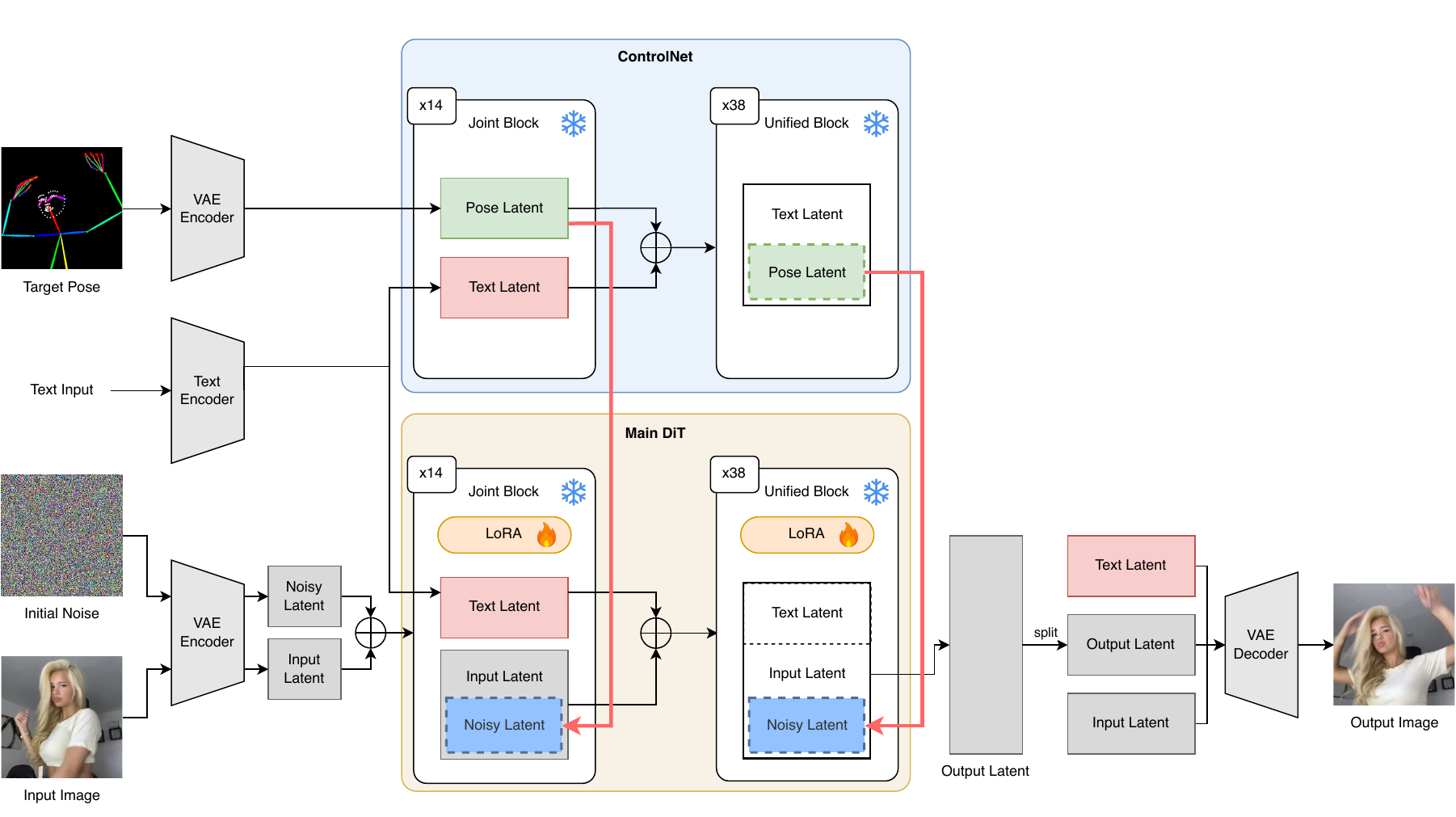}
  \caption{\textbf{Pose-conditioned generator $G$.} $G$ is built on
  FLUX.1~Kontext~\cite{flux2025kontext} paired with a pose
  ControlNet~\cite{zhang2023controlnet}. The target pose is VAE-encoded and read by a
  frozen ControlNet ($14$ joint blocks, $38$ unified blocks), whose pose-aware
  features condition the main DiT. The reference image and initial noise are encoded
  and denoised by the main DiT, into which we inject the trainable
  LoRA~\cite{hu2022lora}; a text encoder supplies the prompt. The output latent is
  VAE-decoded to the subject rendered under the target pose. Backbone weights are
  frozen (snowflake); only the LoRA adapters are trained (flame).}
  \label{fig:generator}
\end{figure*}

\section{Further Implementation Details}
\label{app:impl}

\subsection{Base Model and Generator Architectures}
\label{app:impl:arch}
The base animator $F_{\theta}$ is UniAnimate-DiT~\cite{wang2025unianimatedit}, a
pose-driven human video diffusion transformer built on the Wan2.1-I2V (14B)
backbone~\cite{wan2025}. It operates in the compressed latent space of a frozen
video VAE and conditions on three streams: a reference-appearance stream, in which
the source image is encoded by the frozen VAE encoder and concatenated with the
noisy video latents to preserve high-frequency detail; a target-pose stream, which
injects the driving skeleton as a residual at the latent entry; and dual
cross-attention with dedicated image-token projections for reference alignment. The
backbone uses a single unified 3D self-attention stack that models space and time
jointly. We keep all pre-trained parameters frozen and adapt only a per-subject
low-rank update $\Delta\theta$.

The generator $G$ is built on FLUX.1 Kontext~\cite{flux2025kontext}, a
FLUX.1~\cite{flux2024} variant chosen for its strong reference and prior
preservation, paired with a pose ControlNet~\cite{zhang2023controlnet}
(\cref{fig:generator}). We freeze
the base and train a low-rank LoRA~\cite{hu2022lora} on its DiT attention blocks
with the two-frame reconstruction objective of the main paper, so that $G$ renders
the reference subject under a target pose while preserving identity.

\subsection{Key-pose and frame selection}
\label{app:impl:select}
\noindent\textbf{Which poses are supervised.} We do not generate a frame for every
pose in $\mathcal{P}$; we supervise only a small set of $K$ key poses. The key poses
are chosen by greedy farthest-point sampling in 2D keypoint space, initialized at the
reference pose and adding, at each step, the pose whose minimum keypoint distance to
the already-selected set is largest. This spreads the key poses across the range of
motion and concentrates supervision on the poses farthest from the reference, which
the base animator renders worst.
The keypoint distance is the mean per-joint Euclidean distance between the normalized
2D whole-body keypoints (extracted by DWPose~\cite{yang2023effectivewholebodyposeestimation}) of the two frames.
By default $K{=}5$: the reference pose plus the four farthest poses. For example, on
TikTok sequence~07 the selected key frames are $\{0,\,38,\,53,\,57,\,71\}$ of the
$305$-frame clip.

\noindent\textbf{Which frame is kept.} For each key pose, the reward-guided sampler
(\cref{alg:guided}) produces a single frame: $N$ particles denoise under in-loop DINO
steering toward $x_{\mathrm{ref}}$, and the surviving candidate with the highest
ArcFace identity is kept. There is no post-hoc filtering of a larger pool; selection
is the in-loop steering followed by the final ArcFace pick.
We use $N{=}8$ particles and resample at $|\mathcal{R}|{=}3$ points evenly spaced over
the back half of the $30$-step denoising schedule (start fraction $0.5$; resample
steps $\{18,22,25\}$), injecting churn $\eta{=}0.5$ so cloned winners diverge under the
otherwise-deterministic dynamics. The in-loop steering reward weights a DINOv2
body-similarity term ($w_b{=}1.0$) and a Laplacian sharpness term ($w_\sigma{=}0.5$);
the surviving per-pose candidate is then chosen by maximum ArcFace identity to
$x_{\mathrm{ref}}$.

\subsection{Efficient Adaptation via LoRA}
\label{app:impl:lora}
We inject LoRA~\cite{hu2022lora} into the query, key, value, and output projections
of the attention blocks and the two feed-forward layers ($\mathrm{ffn.0},
\mathrm{ffn.2}$) of every transformer block, with rank $r$ and scaling $\alpha=r$
following standard convention. Freezing the backbone and adapting only these low-rank
factors keeps the trainable parameter count a small fraction of the backbone and the
adaptation fast. Concretely, the animator $F$ uses $r{=}128$ ($\alpha{=}128$) on
$\{q,k,v,o,\mathrm{ffn.0},\mathrm{ffn.2}\}$ of each DiT block ($\approx\!0.61$B
trainable parameters, $\sim\!4\%$ of the frozen $14$B backbone), and the generator $G$
uses $r{=}64$ on its DiT attention, feed-forward, and modulation projections
($\approx\!0.31$B trainable parameters).

\subsection{Training Details}
\label{app:impl:train}
\noindent\textbf{Generator training.} We optimize $G$ with AdamW at a constant
learning rate of $1\times10^{-4}$ for $4{,}000$ steps on a single H100 GPU, on the
in-domain training split of each benchmark. We train at LoRA rank $r{=}64$ with one
(reference, target) frame pair per step and gradient checkpointing; supervision is
real frame pairs, so no photometric augmentation is applied. The cross-frame pairing
itself supplies the pose variation.

\noindent\textbf{Per-subject adaptation.} At test time we synthesize $K$ pseudo
frames, replicate each into a static clip, and adapt $\Delta\theta$ on these clips.
Adaptation is a one-time per-subject cost; once $\Delta\theta$ is obtained,
inference along an arbitrary driving sequence requires no further optimization.
Adaptation uses AdamW at learning rate $1\times10^{-4}$ for $5$ steps by default
(\cref{tab:finetunetime} sweeps $5$--$400$).

\section{Evaluation Metrics}
\label{app:metrics}
We provide the formulations of every metric used in our experiments. Let $I_t$ and
$\hat{I}_t$ denote the ground-truth and generated frames at time $t$, with spatial
size $H\times W$, $C$ channels, and $T$ total frames.

\noindent\textbf{PSNR.} From the per-frame mean squared error
$\mathrm{MSE}=\tfrac{1}{HWC}\sum_{x,y,c}\big(I_t(x,y,c)-\hat{I}_t(x,y,c)\big)^2$,
\begin{equation}
  \mathrm{PSNR}=20\log_{10}\!\left(\frac{255}{\sqrt{\mathrm{MSE}}}\right),
\end{equation}
averaged over frames. Higher is better.

\noindent\textbf{SSIM.} Structural similarity~\cite{SSIM} compares local luminance, contrast,
and structure between $I_t$ and $\hat{I}_t$. We report the mean over all frames and
channels. Higher is better.

\noindent\textbf{LPIPS.} Learned Perceptual Image Patch Similarity~\cite{LPIPS} measures the
distance between deep features of $I_t$ and $\hat{I}_t$ from a pre-trained backbone,
averaged over frames. Lower is better.

\noindent\textbf{FID.} Fr\'echet Inception Distance~\cite{heusel2017gans} compares Gaussian statistics
$(\mu_r,\Sigma_r)$ and $(\mu_f,\Sigma_f)$ of real and generated frame features from
a pre-trained Inception network,
\begin{equation}
  \mathrm{FID}=\lVert\mu_r-\mu_f\rVert_2^2
  +\mathrm{Tr}\!\big(\Sigma_r+\Sigma_f-2(\Sigma_r\Sigma_f)^{1/2}\big).
\end{equation}

\noindent\textbf{FVD.} Fr\'echet Video Distance~\cite{unterthiner2019accurategenerativemodelsvideo}
applies the same form to spatiotemporal features extracted by an Inflated 3D
ConvNet~\cite{Carreira2017QuoVA} from fixed-length video segments, capturing temporal
realism. Lower is better.

\noindent\textbf{FID-VID.} A video-level variant~\cite{balaji2019conditional} that aggregates clip embeddings
into a single descriptor per sequence before computing the Fr\'echet distance,
capturing long-range temporal consistency. Lower is better.

\noindent\textbf{FaceID.} Facial identity preservation is the cosine similarity
between ArcFace~\cite{deng2019arcface} embeddings of the generated and reference
faces, averaged over frames. Higher is better.

\noindent\textbf{BodySIM.} Body-appearance similarity is the cosine similarity
between DINOv2~\cite{oquab2024dinov2} embeddings of the generated and reference
subject crops. Higher is better.

\noindent\textbf{dtSSD.} Temporal consistency is measured on inter-frame
differences: we compare the temporal gradient of the generated video,
$\hat{I}_{t+1}-\hat{I}_t$, against that of the reference video, $I_{t+1}-I_t$, and
report the mean squared deviation over time. Lower indicates motion closer to the
ground truth without added flicker.

\noindent\textbf{PoseAcc.} Pose-following accuracy compares the skeleton extracted
from the generated frame against the driving skeleton, reporting keypoint
agreement. Poses are extracted with DWPose~\cite{yang2023effectivewholebodyposeestimation} from both the
generated and the ground-truth frame, and agreement is the percentage of correct
keypoints (PCK@$0.2$): a predicted joint is counted correct when its distance to the
corresponding ground-truth joint is within $0.2$ of the person bounding-box size.
Higher indicates closer adherence to the target motion.

\noindent We follow the overall evaluation protocol of
RealisDance-DiT~\cite{zhou2025realisdancedit} (see \cref{app:data:proto}).

\subsection{Identity under independent recognizers}
\label{app:metrics:indep}
FaceID is ArcFace cosine and BodySIM is DINOv2 cosine, and those same two
networks drive the pipeline: candidates are selected by ArcFace similarity to the
reference and sampling is steered by DINOv2 body similarity. To verify that an identity gain measured with them reflects identity rather than the
selection reward, \Cref{tab:indepidentity} re-scores the same rendered videos with
recognizers that are not used at any stage of the pipeline. Two controls ensure a fair comparison: faces are
detected and aligned once and every recognizer embeds the identical crop, so
detector disagreement cannot appear as a difference between recognizers, and the
scored set is fixed across arms, so no arm benefits from being evaluated on the
frames on which its own faces are detectable.

The gain persists and is larger under FaceNet512 ($+0.059$) and SFace
($+0.067$) than under ArcFace ($+0.035$). Exploiting ArcFace-specific artifacts would
produce the opposite ordering. We also run the test in reverse, rebuilding the
per-subject dataset by ranking the same candidate pool with FaceNet512 instead of
ArcFace (Ours-FN in \cref{tab:indepidentity}). The two recognizers disagree on
$7$ of $12$ selections, so the two training sets differ substantially. Scored with
ArcFace, which it never optimized, Ours-FN gains $+0.034$ over the base. Although Ours-FN optimizes FaceNet512, it does not lead on FaceNet512: selection does
not inflate the score of the network doing the selecting.
Appearance evidence that does not depend on either reward is provided by LPIPS, which is computed on VGG features and
improves substantially ($0.177$ vs $0.283$, \cref{tab:gtuning}). PSNR, SSIM and FVD are likewise
reward-independent and all improve (\cref{tab:main}).

\begin{table}[t]
  \centering
  \caption{\textbf{Identity measured by recognizers not used by the pipeline.}
  Pseudo-frames are selected with ArcFace, so identity is re-scored with FaceNet512 and
  SFace, which are not used in selection. Faces are detected once and every recognizer embeds the
  identical crop. A frame is scored only when a face is found in the ground truth and in
  both arms ($33$ pairs for base vs ours, $30$ for the three-arm comparison, trim regime,
  3 TikTok sequences). Ours-FN selects with FaceNet512 instead of ArcFace.}
  \label{tab:indepidentity}
  \scriptsize
  \setlength{\tabcolsep}{5pt}
  \begin{adjustbox}{max width=\columnwidth}
\begin{tabular}{@{}lcccc@{}}
    \toprule
    Metric & Base & Ours & Ours-FN & Not used by \\
    \midrule
    ArcFace    & 0.690 & 0.731 & \textbf{0.724} & Ours-FN \\
    FaceNet512 & 0.722 & \textbf{0.786} & 0.778 & Ours \\
    SFace      & 0.626 & \textbf{0.704} & \textbf{0.704} & both \\
    \bottomrule
  \end{tabular}
  \end{adjustbox}
\end{table}

\subsection{Human preference study}
\label{app:metrics:human}
Every metric above is automatic, and two of them (FaceID, BodySIM) share their
networks with the reward used to select pseudo-frames (\cref{app:metrics:indep}).
We therefore also conducted a human study, which depends on neither.

Thirty-four respondents rated clips from each method on a 1--5 scale along four axes:
human appearance alignment, pose alignment, background alignment, and overall
quality. Clips were drawn from the TikTok test split, method identities were hidden,
and the presentation order of the four methods was shuffled per clip so that no
position was systematically associated with one method. Each respondent saw the
reference image alongside the clip, since appearance alignment is only meaningful
relative to it. This yields $180$ ratings per method per axis.

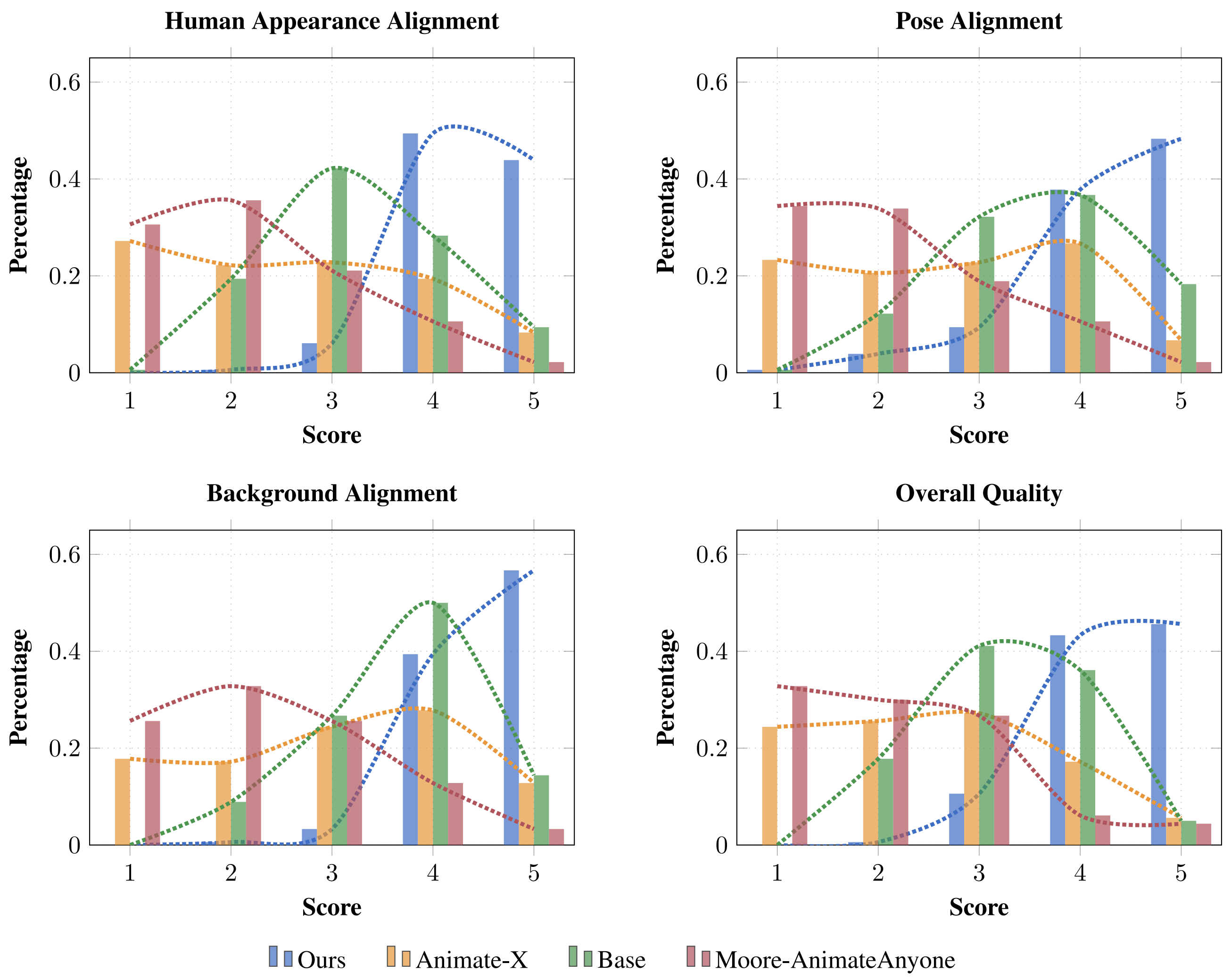
\begin{figure*}[t!]
    \centering
    \colorlet{oursColor}{surveyOurs}
    \colorlet{animateXColor}{surveyAnimateX}
    \colorlet{baseMethodColor}{surveyBase}
    \colorlet{mooreColor}{surveyMoore}
    \pgfplotsset{
        every axis/.append style={
            ybar,
            bar width=6pt,
            width=\linewidth,%
            height=6cm,
            ymin=0, ymax=0.65,
            xlabel={\textbf{Score}},
            ylabel={\textbf{Percentage}},
            xtick={1,2,3,4,5},
            grid=major,
            grid style={dotted, gray!40},
            ylabel near ticks,
            xlabel near ticks
        }
    }

    \begin{subfigure}[t]{0.48\textwidth}
        \centering
        \begin{tikzpicture}
            \begin{axis}[title={\textbf{Human Appearance Alignment}}]

            \addplot[fill=oursColor, draw=none, opacity=0.7, bar shift=-9pt] coordinates {(1,0.000) (2,0.006) (3,0.061) (4,0.494) (5,0.439)};
            \addplot[draw=oursColor, ultra thick, densely dotted, smooth, tension=0.5, bar shift=-9pt, forget plot] coordinates {(1,0.000) (2,0.006) (3,0.061) (4,0.494) (5,0.439)};

            \addplot[fill=animateXColor, draw=none, opacity=0.7, bar shift=-3pt] coordinates {(1,0.272) (2,0.222) (3,0.228) (4,0.194) (5,0.083)};
            \addplot[draw=animateXColor, ultra thick, densely dotted, smooth, tension=0.5, bar shift=-3pt, forget plot] coordinates {(1,0.272) (2,0.222) (3,0.228) (4,0.194) (5,0.083)};

            \addplot[fill=baseMethodColor, draw=none, opacity=0.7, bar shift=+3pt] coordinates {(1,0.006) (2,0.194) (3,0.422) (4,0.283) (5,0.094)};
            \addplot[draw=baseMethodColor, ultra thick, densely dotted, smooth, tension=0.5, bar shift=+3pt, forget plot] coordinates {(1,0.006) (2,0.194) (3,0.422) (4,0.283) (5,0.094)};

            \addplot[fill=mooreColor, draw=none, opacity=0.7, bar shift=+9pt] coordinates {(1,0.306) (2,0.356) (3,0.211) (4,0.106) (5,0.022)};
            \addplot[draw=mooreColor, ultra thick, densely dotted, smooth, tension=0.5, bar shift=+9pt, forget plot] coordinates {(1,0.306) (2,0.356) (3,0.211) (4,0.106) (5,0.022)};

            \end{axis}
        \end{tikzpicture}
        \label{fig:cat_a}
    \end{subfigure}%
    \hfill%
    \begin{subfigure}[t]{0.48\textwidth}
        \centering
        \begin{tikzpicture}
            \begin{axis}[title={\textbf{Pose Alignment}}]

            \addplot[fill=oursColor, draw=none, opacity=0.7, bar shift=-9pt] coordinates {(1,0.006) (2,0.039) (3,0.094) (4,0.378) (5,0.483)};
            \addplot[draw=oursColor, ultra thick, densely dotted, smooth, tension=0.5, bar shift=-9pt, forget plot] coordinates {(1,0.006) (2,0.039) (3,0.094) (4,0.378) (5,0.483)};

            \addplot[fill=animateXColor, draw=none, opacity=0.7, bar shift=-3pt] coordinates {(1,0.233) (2,0.206) (3,0.228) (4,0.267) (5,0.067)};
            \addplot[draw=animateXColor, ultra thick, densely dotted, smooth, tension=0.5, bar shift=-3pt, forget plot] coordinates {(1,0.233) (2,0.206) (3,0.228) (4,0.267) (5,0.067)};

            \addplot[fill=baseMethodColor, draw=none, opacity=0.7, bar shift=+3pt] coordinates {(1,0.006) (2,0.122) (3,0.322) (4,0.367) (5,0.183)};
            \addplot[draw=baseMethodColor, ultra thick, densely dotted, smooth, tension=0.5, bar shift=+3pt, forget plot] coordinates {(1,0.006) (2,0.122) (3,0.322) (4,0.367) (5,0.183)};

            \addplot[fill=mooreColor, draw=none, opacity=0.7, bar shift=+9pt] coordinates {(1,0.344) (2,0.339) (3,0.189) (4,0.106) (5,0.022)};
            \addplot[draw=mooreColor, ultra thick, densely dotted, smooth, tension=0.5, bar shift=+9pt, forget plot] coordinates {(1,0.344) (2,0.339) (3,0.189) (4,0.106) (5,0.022)};

            \end{axis}
        \end{tikzpicture}
        \label{fig:cat_b}
    \end{subfigure}

    \vspace{0.3cm}%

    \begin{subfigure}[t]{0.48\textwidth}
        \centering
        \begin{tikzpicture}
            \begin{axis}[title={\textbf{Background Alignment}}]

            \addplot[fill=oursColor, draw=none, opacity=0.7, bar shift=-9pt] coordinates {(1,0.000) (2,0.006) (3,0.033) (4,0.394) (5,0.567)};
            \addplot[draw=oursColor, ultra thick, densely dotted, smooth, tension=0.5, bar shift=-9pt, forget plot] coordinates {(1,0.000) (2,0.006) (3,0.033) (4,0.394) (5,0.567)};

            \addplot[fill=animateXColor, draw=none, opacity=0.7, bar shift=-3pt] coordinates {(1,0.178) (2,0.172) (3,0.244) (4,0.278) (5,0.128)};
            \addplot[draw=animateXColor, ultra thick, densely dotted, smooth, tension=0.5, bar shift=-3pt, forget plot] coordinates {(1,0.178) (2,0.172) (3,0.244) (4,0.278) (5,0.128)};

            \addplot[fill=baseMethodColor, draw=none, opacity=0.7, bar shift=+3pt] coordinates {(1,0.000) (2,0.089) (3,0.267) (4,0.500) (5,0.144)};
            \addplot[draw=baseMethodColor, ultra thick, densely dotted, smooth, tension=0.5, bar shift=+3pt, forget plot] coordinates {(1,0.000) (2,0.089) (3,0.267) (4,0.500) (5,0.144)};

            \addplot[fill=mooreColor, draw=none, opacity=0.7, bar shift=+9pt] coordinates {(1,0.256) (2,0.328) (3,0.256) (4,0.128) (5,0.033)};
            \addplot[draw=mooreColor, ultra thick, densely dotted, smooth, tension=0.5, bar shift=+9pt, forget plot] coordinates {(1,0.256) (2,0.328) (3,0.256) (4,0.128) (5,0.033)};

            \end{axis}
        \end{tikzpicture}
        \label{fig:cat_c}
    \end{subfigure}%
    \hfill
    \begin{subfigure}[t]{0.48\textwidth}
        \centering
        \begin{tikzpicture}
            \begin{axis}[title={\textbf{Overall Quality}}]

            \addplot[fill=oursColor, draw=none, opacity=0.7, bar shift=-9pt] coordinates {(1,0.000) (2,0.006) (3,0.106) (4,0.433) (5,0.456)};
            \addplot[draw=oursColor, ultra thick, densely dotted, smooth, tension=0.5, bar shift=-9pt, forget plot] coordinates {(1,0.000) (2,0.006) (3,0.106) (4,0.433) (5,0.456)};

            \addplot[fill=animateXColor, draw=none, opacity=0.7, bar shift=-3pt] coordinates {(1,0.244) (2,0.256) (3,0.272) (4,0.172) (5,0.056)};
            \addplot[draw=animateXColor, ultra thick, densely dotted, smooth, tension=0.5, bar shift=-3pt, forget plot] coordinates {(1,0.244) (2,0.256) (3,0.272) (4,0.172) (5,0.056)};

            \addplot[fill=baseMethodColor, draw=none, opacity=0.7, bar shift=+3pt] coordinates {(1,0.000) (2,0.178) (3,0.411) (4,0.361) (5,0.050)};
            \addplot[draw=baseMethodColor, ultra thick, densely dotted, smooth, tension=0.5, bar shift=+3pt, forget plot] coordinates {(1,0.000) (2,0.178) (3,0.411) (4,0.361) (5,0.050)};

            \addplot[fill=mooreColor, draw=none, opacity=0.7, bar shift=+9pt] coordinates {(1,0.328) (2,0.300) (3,0.267) (4,0.061) (5,0.044)};
            \addplot[draw=mooreColor, ultra thick, densely dotted, smooth, tension=0.5, bar shift=+9pt, forget plot] coordinates {(1,0.328) (2,0.300) (3,0.267) (4,0.061) (5,0.044)};

            \end{axis}
        \end{tikzpicture}
        \label{fig:cat_d}
    \end{subfigure}

    \vspace{0.2cm}

    \begin{tikzpicture}
        \begin{axis}[
            hide axis, scale only axis, width=1mm, height=1mm,
            xmin=0, xmax=1, ymin=0, ymax=1,
            legend style={
                at={(0.5,0.5)}, anchor=center, legend columns=-1,
                draw=none, fill=none,
                /tikz/every even column/.append style={column sep=0.5cm}
            }
        ]
        \addlegendimage{ybar, ybar legend, fill=oursColor, draw=none, opacity=0.7}
        \addlegendentry{Ours}
        \addlegendimage{ybar, ybar legend, fill=animateXColor, draw=none, opacity=0.7}
        \addlegendentry{Animate-X}
        \addlegendimage{ybar, ybar legend, fill=baseMethodColor, draw=none, opacity=0.7}
        \addlegendentry{Base}
        \addlegendimage{ybar, ybar legend, fill=mooreColor, draw=none, opacity=0.7}
        \addlegendentry{Moore-AnimateAnyone}
        \end{axis}
    \end{tikzpicture}

    \caption{\textbf{Human preference study.} Distribution of 1--5 ratings for
    ours, Animate-X, the unadapted base, and Moore-AnimateAnyone, on four axes.
    Thirty-four respondents, $180$ ratings per bar group.
    Ours places $93\%$ of appearance-alignment ratings at 4--5, against $38\%$ for the
    base model; the ordering is the same on all four axes. Protocol in
    \cref{app:metrics:human}.}
    \label{fig:survey_results}
\end{figure*}

\Cref{fig:survey_results} shows the resulting distributions. Ours places $93\%$
of appearance-alignment ratings at 4--5 against $38\%$ for the unadapted base, and
leads on all four axes. The base model in turn leads both prior animators, which is
consistent with the automatic metrics in \cref{tab:main}, and corroborates the
perceptual metrics with judgments that depend on neither reward.

\section{Further Ablation}
\label{app:ablation}

\subsection{Additional Quantitative Results}
\label{app:ablation:quant}
\noindent\textbf{Quality versus fine-tuning time.}
\label{app:time}
\Cref{tab:finetunetime} studies how reconstruction, identity, and video quality
scale with the number of per-subject fine-tuning steps and the corresponding
wall-clock time on a single H100. Quality saturates by five steps, our default, and is flat over the rest of the
sweep: five steps already match or exceed every larger budget on reconstruction and
video quality, since the static clip presents the same target at every step and the
low-rank update converges to it within a few updates. Longer budgets improve only
identity, and not monotonically, so we adapt for five steps, at $5.8$\,s per subject.
This sweep varies the step budget only. Peak GPU memory, the hardware
requirement, and the cost of the other pipeline stages are reported separately in
\cref{app:cost}.

\begin{table}[tbp]
  \centering
  \caption{\textbf{Quality versus fine-tuning time.} Per-subject adaptation quality at
  different numbers of fine-tuning steps and the time they take on a single H100
  ($K{=}5$, TikTok). Quality saturates within the first five steps, our default. Longer budgets improve
  only identity, and not monotonically.}
  \label{tab:finetunetime}
  \scriptsize
  \setlength{\tabcolsep}{3pt}
  \begin{adjustbox}{max width=\columnwidth}
\begin{tabular}{@{}lcccccccc@{}}
    \toprule
    Steps & Time (s) & PSNR$\uparrow$ & SSIM$\uparrow$ & LPIPS$\downarrow$ &
    FaceID$\uparrow$ & BodySIM$\uparrow$ & FVD$\downarrow$ & dtSSD$\downarrow$ \\
    \midrule
    \textbf{5 (ours)} & \textbf{5.8} & \textbf{22.89} &
      \textbf{0.853} & \textbf{0.166} & 0.772 & 0.939 &
      \textbf{289.4} & \textbf{212.0} \\
    25  & 29  & 22.71 & 0.848 & 0.169 & 0.764 & 0.935 & 352.9 & 218.3 \\
    50  & 58  & 22.26 & 0.838 & 0.181 & 0.820 & 0.931 & 399.4 & 219.0 \\
    100 & 116 & 22.72 & 0.850 & 0.168 & 0.768 & 0.937 & 341.5 & 213.9 \\
    200 & 232 & 22.85 & 0.850 & 0.167 & 0.762 & \textbf{0.941} & 291.7 & 212.3 \\
    400 & 464 & 22.21 & 0.840 & 0.181 & \textbf{0.825} & 0.927 & 380.5 & 217.4 \\
    \bottomrule
  \end{tabular}
  \end{adjustbox}
\end{table}

\noindent\textbf{Per-subject LoRA rank.}
\Cref{tab:lorarank} sweeps the rank $r$ of the per-subject update on a tuning subset.
Quality improves with adapter capacity up to $r{=}128$, our default. At $r{=}256$ the
adapter over-fits the handful of pseudo-frames and quality regresses, marking $128$ as
a clear optimum.

\begin{table}[tbp]
  \centering
  \caption{\textbf{Per-subject LoRA rank $r$} (on a tuning subset). Adaptation quality
  as the rank of the per-subject update varies. Quality improves with capacity up to
  $r{=}128$, our default; beyond it ($r{=}256$) the adapter over-fits the few
  pseudo-frames and quality regresses, marking a clear optimum. Absolute scores differ
  from the main tables because this sweep runs on a tuning subset.}
  \label{tab:lorarank}
  \scriptsize
  \setlength{\tabcolsep}{2pt}
  \renewcommand{\arraystretch}{1.5}
  \resizebox{\linewidth}{!}{%
  \begin{tabular}{@{}lccccccc@{}}
    \toprule
    $r$ & PSNR$\uparrow$ & SSIM$\uparrow$ & LPIPS$\downarrow$ &
    FaceID$\uparrow$ & BodySIM$\uparrow$ & FVD$\downarrow$ & dtSSD$\downarrow$ \\
    \midrule
    $4$   & 20.12 & 0.820 & 0.232 & 0.558 & 0.858 & 176.4 & 410.2 \\
    $8$   & 20.61 & 0.829 & 0.224 & 0.581 & 0.866 & 161.2 & 397.5 \\
    $16$  & 20.94 & 0.837 & 0.216 & 0.597 & 0.872 & 150.1 & 384.0 \\
    $32$  & 21.08 & 0.841 & 0.211 & 0.606 & 0.877 & 143.6 & 379.1 \\
    $64$  & 21.31 & 0.846 & 0.206 & 0.622 & 0.883 & 136.2 & \textbf{363.9} \\
    $\mathbf{128}$ & \textbf{21.49} & \textbf{0.852} & \textbf{0.201} &
    \textbf{0.635} & \textbf{0.890} & \textbf{129.8} & 365.1 \\
    $256$ & 21.22 & 0.844 & 0.212 & 0.615 & 0.881 & 142.7 & 381.4 \\
    \bottomrule
  \end{tabular}}
\end{table}

\subsection{Generator and Reward-Guided Sampling}
\label{app:ablation:gen}
Our pipeline contains a second trained model, the pose-conditioned generator $G$,
and a reference-only reward-guided sampler. We ablate their components here.

\noindent\textbf{How much domain tuning does $G$ need?} $G$'s LoRA is trained on
the in-domain split of each benchmark, so the pipeline assumes access to a
human-video corpus before test time, even though no video of the \emph{test subject}
is ever used. \Cref{tab:gtuning} quantifies the effect of that assumption by
sweeping the amount of tuning $G$ receives, from none to our default. Two conclusions follow. First, the dependence is present but bounded: with an entirely untuned
generator the adapted model still improves over the base on every metric
($21.88$ vs $20.33$ PSNR, $0.184$ vs $0.283$ LPIPS). Second, the benefit of tuning depends on the metric.
Pixel and perceptual quality saturate early: $500$ steps, one eighth of the
default, already recovers $95\%$ of the PSNR gain over base, and LPIPS, SSIM and
BodySIM are flat from there on. Identity does not: FaceID continues to increase to
$4{,}000$ ($0.699 \rightarrow 0.714 \rightarrow 0.720 \rightarrow 0.764$), so $500$
steps recovers only $66\%$ of that gain. Domain tuning of $G$ therefore primarily improves identity fidelity, as expected of a LoRA that teaches $G$ the
appearance statistics of the domain. Reconstruction quality requires only light tuning on a human-video corpus rather than close matching to the benchmark. Consistent with this, the $G$ used for
Unseen100-Val is tuned on TikTok yet leads on in-the-wild subjects
(\cref{tab:crossdomain}). The gains concentrate on the sequence where the base model
performs worst, where PSNR rises from $12.12$ (base) to $17.39$ (untuned $G$) to
$20.50$ (default). On sequences where the base already performs well, the tuning levels are
indistinguishable.

\begin{table}[t]
  \centering
  \caption{\textbf{Sensitivity to the generator's domain tuning.} $G$'s LoRA receives
  from no tuning (off-the-shelf FLUX.1 Kontext with the pose ControlNet) to our default
  of $4{,}000$ steps, with the rest of the pipeline fixed. \emph{Teacher} is the fidelity
  of the selected pseudo-frame to the real frame at that pose, and \emph{downstream} is
  the adapted model's video quality (trim, 3 TikTok sequences).}
  \label{tab:gtuning}
  \scriptsize
  \setlength{\tabcolsep}{4pt}
  \begin{adjustbox}{max width=\columnwidth}
\begin{tabular}{@{}lcccccc@{}}
    \toprule
    & \multicolumn{2}{c}{Teacher (pseudo-frame)} & \multicolumn{4}{c}{Downstream (video)} \\
    \cmidrule(lr){2-3}\cmidrule(l){4-7}
    $G$ tuning & PSNR$\uparrow$ & BodySim$\uparrow$ &
    PSNR$\uparrow$ & LPIPS$\downarrow$ & FaceID$\uparrow$ & BodySim$\uparrow$ \\
    \midrule
    Base (no adaptation) & -- & -- &
      20.33 & 0.283 & 0.619 & 0.906 \\
    \addlinespace[2pt]
    None (off-the-shelf) & 17.85 & 0.862 &
      21.88 & 0.184 & 0.699 & 0.928 \\
    $500$ steps          & 17.89 & 0.857 &
      22.75 & 0.176 & 0.714 & 0.928 \\
    $1{,}000$ steps      & 18.54 & 0.873 &
      22.71 & 0.176 & 0.720 & 0.929 \\
    $4{,}000$ steps (ours) & 19.54 & 0.893 &
      22.89 & 0.177 & \textbf{0.764} & 0.927 \\
    \bottomrule
  \end{tabular}
  \end{adjustbox}
\end{table}

\noindent\textbf{Particle count $N$.} \Cref{tab:particlesweep} sweeps the number of
particles $N$ in the reward-guided sampler, reporting the fidelity of the selected
pseudo-frame to the real frame. Frame quality is already largely saturated by our
default $N{=}8$, while per-frame sampling time grows roughly linearly, so larger
populations cost compute without further gain. The value of in-loop steering over
best-of-$N$ selection at a matched budget is established separately in
\cref{tab:rewardablation}.

\begin{table}[t]
  \centering
  \caption{\textbf{Particle count $N$.} Selected pseudo-ground-truth frame fidelity to the real
  frame (mean over the $K$ supervised poses $\times$ 3 sequences) and per-pseudo-frame
  generation time, as the number of reward-guided particles varies. Frame quality is
  largely saturated already by our default $N{=}8$, while sampling time grows roughly
  linearly, so larger populations cost compute without further gain. Time is the
  per-pseudo-frame sampling cost on a single H100.}
  \label{tab:particlesweep}
  \scriptsize
  \setlength{\tabcolsep}{5pt}
  \begin{adjustbox}{max width=\columnwidth}
\begin{tabular}{@{}lcccccc@{}}
    \toprule
    $N$ & Time (s) & PSNR$\uparrow$ & SSIM$\uparrow$ & LPIPS$\downarrow$ & FaceID$\uparrow$ & BodySIM$\uparrow$ \\
    \midrule
    $2$        & 28.1  & 19.71 & 0.824 & 0.232 & 0.718 & 0.891 \\
    $4$        & 41.4  & 19.36 & 0.814 & 0.239 & 0.719 & 0.901 \\
    $8$ (ours) & 69.1  & 19.51 & 0.815 & 0.235 & 0.707 & 0.890 \\
    $16$       & 134.5 & 19.98 & 0.827 & 0.226 & 0.718 & 0.886 \\
    \bottomrule
  \end{tabular}
  \end{adjustbox}
\end{table}

\noindent\textbf{Resampling schedule.} \Cref{tab:resamplesched} varies the denoising
window over which particles are resampled and the churn strength $\eta$. Pseudo-frame
fidelity is largely insensitive to the schedule. We default to resampling only the
back half with churn for its downstream stability, since steering early, high-noise
particles collapses diversity before subject structure emerges and the churn step
keeps duplicated high-reward particles from remaining identical under the
deterministic dynamics.

\begin{table}[t]
  \centering
  \caption{\textbf{Resampling schedule.} Selected pseudo-ground-truth frame fidelity to the real
  frame as a function of the denoising window over which particles are resampled and
  the churn strength $\eta$ that re-noises duplicated particles ($N{=}8$; mean over the
  $K$ supervised poses $\times$ 3 sequences). Pseudo-ground-truth frame quality is largely
  insensitive to the schedule; the back-half-with-churn default is used for its
  downstream stability.}
  \label{tab:resamplesched}
  \scriptsize
  \setlength{\tabcolsep}{5pt}
  \begin{adjustbox}{max width=\columnwidth}
\begin{tabular}{@{}lccccc@{}}
    \toprule
    Schedule & PSNR$\uparrow$ & SSIM$\uparrow$ & LPIPS$\downarrow$ & FaceID$\uparrow$ & BodySIM$\uparrow$ \\
    \midrule
    Full (all steps)                 & 19.43 & 0.817 & 0.234 & 0.732 & 0.892 \\
    Last quarter                     & 19.61 & 0.816 & 0.235 & 0.751 & 0.893 \\
    Back half (ours)                 & 19.51 & 0.815 & 0.235 & 0.707 & 0.890 \\
    Back half, no churn ($\eta{=}0$) & 19.46 & 0.812 & 0.239 & 0.714 & 0.888 \\
    \bottomrule
  \end{tabular}
  \end{adjustbox}
\end{table}

\noindent\textbf{Key-pose selection.} \Cref{tab:keypose} compares strategies for
choosing the $K$ supervised poses from the driving sequence. At fixed $K$,
farthest-point sampling in keypoint space (ours), random, and uniform-in-time
selection are within noise of one another, indicating the gain comes from supervising
diverse poses at all rather than from the specific selection rule. We use
farthest-point sampling as a principled default.

\begin{table}[t]
  \centering
  \caption{\textbf{Key-pose selection.} How the $K$ supervised poses are chosen from
  the driving sequence, at fixed $K$: farthest-point sampling in keypoint space (ours)
  vs.\ random and uniform-in-time selection, each adapted with an identical step budget and evaluated on
  the generated video (mean over 3 sequences). On this subset the three strategies are
  within noise of one another, indicating the gain comes from supervising diverse poses
  at all rather than from the specific selection rule.}
  \label{tab:keypose}
  \scriptsize
  \setlength{\tabcolsep}{6pt}
  \begin{adjustbox}{max width=\columnwidth}
\begin{tabular}{@{}lcccc@{}}
    \toprule
    Selection & FaceID$\uparrow$ & BodySIM$\uparrow$ & FVD$\downarrow$ & dtSSD$\downarrow$ \\
    \midrule
    Random                & 0.783 & 0.941 & 319.1 & 216.6 \\
    Uniform (in time)     & 0.776 & 0.941 & 293.4 & 215.5 \\
    Farthest-point (ours) & 0.768 & 0.937 & 341.5 & 213.9 \\
    \bottomrule
  \end{tabular}
  \end{adjustbox}
\end{table}

\section{Dataset Curation \& Evaluation Protocol}
\label{app:data}

\subsection{Datasets}
\label{app:data:sets}
For in-domain training and evaluation we use the
TikTok~\cite{jafarian2021tiktok} and UBC Fashion~\cite{zablotskaia2019dwnet}
benchmarks with their standard train and test splits. To probe cross-domain
transfer, we additionally evaluate zero-shot on Unseen100-Val, a set of $100$
in-the-wild videos spanning diverse subjects, motion, lighting, and backgrounds.

\noindent\textbf{Motivation.}
Both standard benchmarks are limited for evaluating modern video diffusion models. The
TikTok test set contains only $10$ videos, too few for fine-grained comparison, while
UBC Fashion, though larger, consists of simple catwalk sequences with uniform lighting
and minimal background variation, so scores saturate across methods
(\cref{tab:main}). Neither covers the diversity of real-world deployment, including
varied skin tones, challenging lighting, and non-photorealistic subjects. We therefore
curate \textbf{Unseen100-Val}, a $100$-video zero-shot evaluation set designed to
(a)~better represent the visual diversity of real-world data and (b)~stress-test
models in challenging settings they were not trained on, such as difficult lighting,
complex backgrounds, and animated characters.

\noindent\textbf{Collection and eligibility.}
Videos are sourced from a mix of TikTok and stock footage. Each clip features a single
person or character performing a dancing motion, with no heavy occlusion of any body
part (slight tolerance is allowed for an arm briefly disappearing during fast motion),
and the pose skeleton must remain fully visible throughout, ensuring reliable pose
extraction for both adaptation and evaluation.

\noindent\textbf{Composition and diversity.}
\Cref{tab:unseen100} summarizes the dataset composition across several axes; each
attribute group independently covers all $100$ videos. We deliberately balance real
and animated subjects and ensure broad coverage of appearance, motion complexity, and
environmental conditions. Among the $80$ real-human videos, skin tones span East
Asian, South Asian, Caucasian, and darker-skinned appearances to promote evaluation
fairness; the $20$ animated videos split evenly between 2D (anime, cartoon) and 3D
(rendered, stylized) characters, testing generalization beyond photorealistic
subjects.

\begin{table}[tbp]
  \centering
  \caption{\textbf{Composition of the Unseen100-Val dataset.} Each attribute group
  independently covers all $100$ videos; rows within a group sum to $100$ (or to $80$
  for appearance and setting, which apply only to the real-human videos).}
  \label{tab:unseen100}
  \scriptsize
  \setlength{\tabcolsep}{6pt}
  \begin{adjustbox}{max width=\columnwidth}
\begin{tabular}{@{}llc@{}}
    \toprule
    Attribute & Category & Count \\
    \midrule
    Subject type     & Real            & 80 \\
                     & Animated -- 2D  & 10 \\
                     & Animated -- 3D  & 10 \\
    \midrule
    Setting          & Indoor          & 54 \\
                     & Outdoor         & 26 \\
    \midrule
    Appearance       & East Asian      & 32 \\
                     & South Asian     & 06 \\
                     & Caucasian       & 24 \\
                     & Darker-skinned  & 18 \\
    \midrule
    Motion intensity & Slow / moderate & 48 \\
                     & Fast / complex  & 52 \\
    \midrule
    Lighting         & Standard        & 61 \\
                     & Challenging     & 39 \\
    \midrule
    Camera           & Static          & 71 \\
                     & Slight shift    & 29 \\
    \bottomrule
  \end{tabular}
  \end{adjustbox}
\end{table}

\noindent\textbf{Preprocessing and statistics.}
All clips are center-cropped and resized to a common resolution, with driving poses
extracted by a 2D keypoint detector consistent with our training pipeline. The
detector is DWPose~\cite{yang2023effectivewholebodyposeestimation}, the same whole-body extractor used to render
the training pose maps; keypoints are extracted at the clip's native resolution and
rendered as skeleton maps at the model input resolution of $832\times480$.
The average clip duration is $14.4$ seconds ($\pm 3.2$\,s), ranging from $8$ to
$24$\,s. At $100$ videos, Unseen100-Val matches the test-set size of UBC Fashion while
being $10\times$ larger than TikTok's test split; because it targets zero-shot
cross-domain evaluation rather than statistical power over a narrow domain, $100$
diverse videos provide sufficient coverage to surface meaningful failure modes and
performance differences across methods.

\subsection{Evaluation Protocol}
\label{app:data:proto}
We follow the evaluation settings of RealisDance-DiT~\cite{zhou2025realisdancedit}.
All generated videos are center-cropped and resized to a common resolution for metric
computation, ensuring a fair comparison across methods that train at different native
resolutions. For competing methods we use officially released checkpoints and
evaluation code where available, and directly report published numbers where
applicable.
Our base animator and adapted model are sampled with the Wan2.1 flow-matching solver
for $50$ steps at classifier-free guidance scale $1.0$, and metrics are computed at a
common $832\times480$ resolution (image metrics on the trimmed window, video metrics
over full clips). Pseudo-frames are generated with the fixed prompt ``\textit{The person is dancing.}''
We do not re-train any competing method: their scores come from official
checkpoints and evaluation code, or from the published papers where applicable.
All baselines in \cref{tab:main}, including the UniAnimate-DiT base, are quoted from
MTVCraft~\cite{ding2025mtvcraft}; the base rows of \cref{tab:buildup,tab:source} re-evaluate it under
our protocol, so its numbers there differ.

\section{Additional Qualitative Results}
\label{app:qual}
\begin{figure*}[p]
  \centering
  \includegraphics[width=\linewidth]{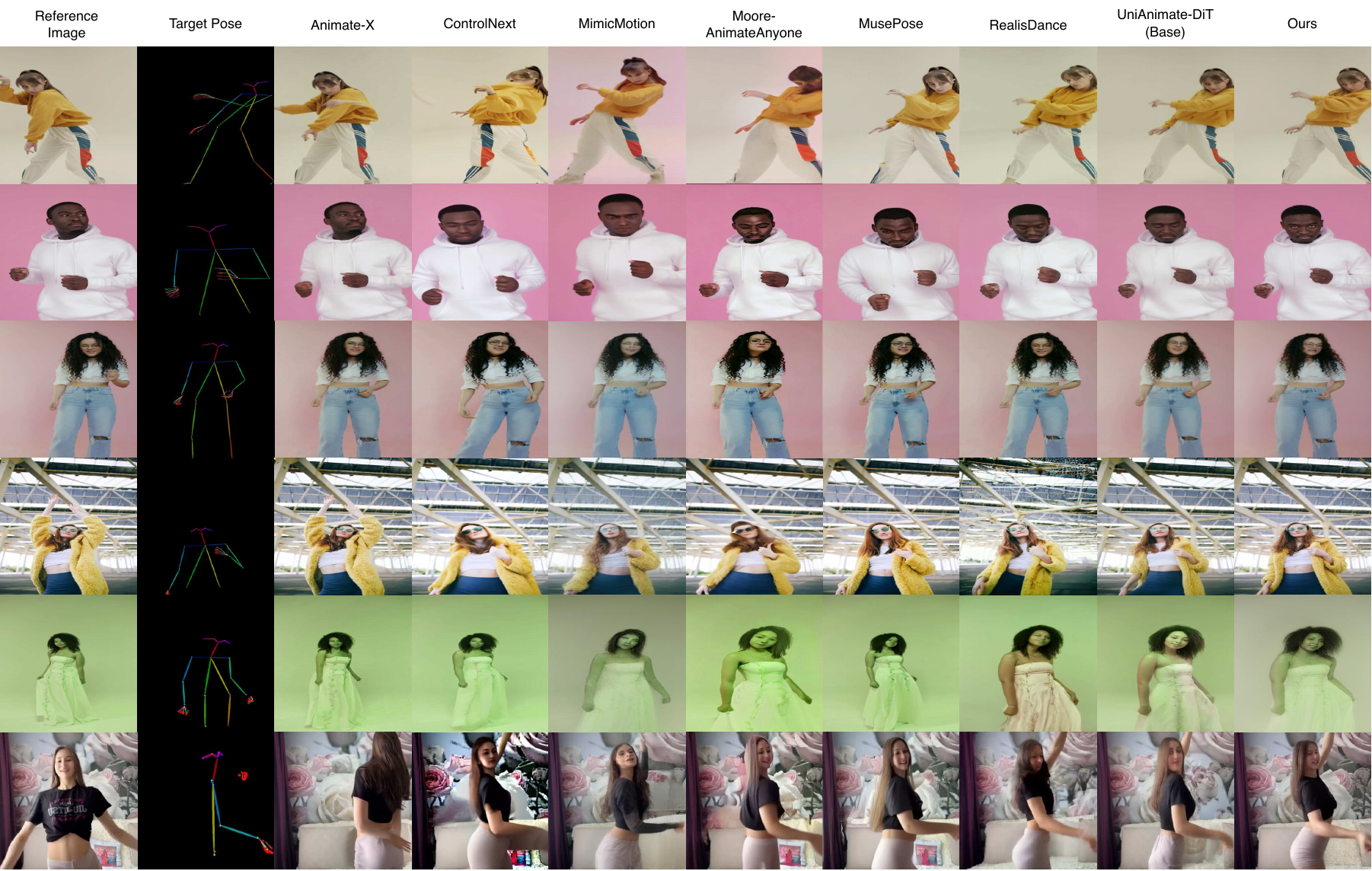}
  \caption{\textbf{In-Domain Comparisons: Diverse Motions and Backgrounds.}
  Examples illustrating variations in background, motion complexity, and clothing
  style. Our approach consistently preserves holistic appearance and pose alignment
  while producing sharper, more realistic details than competing methods.}
  \label{fig:qual-supp1}
\end{figure*}

\begin{figure*}[p]
  \centering
  \includegraphics[width=\linewidth]{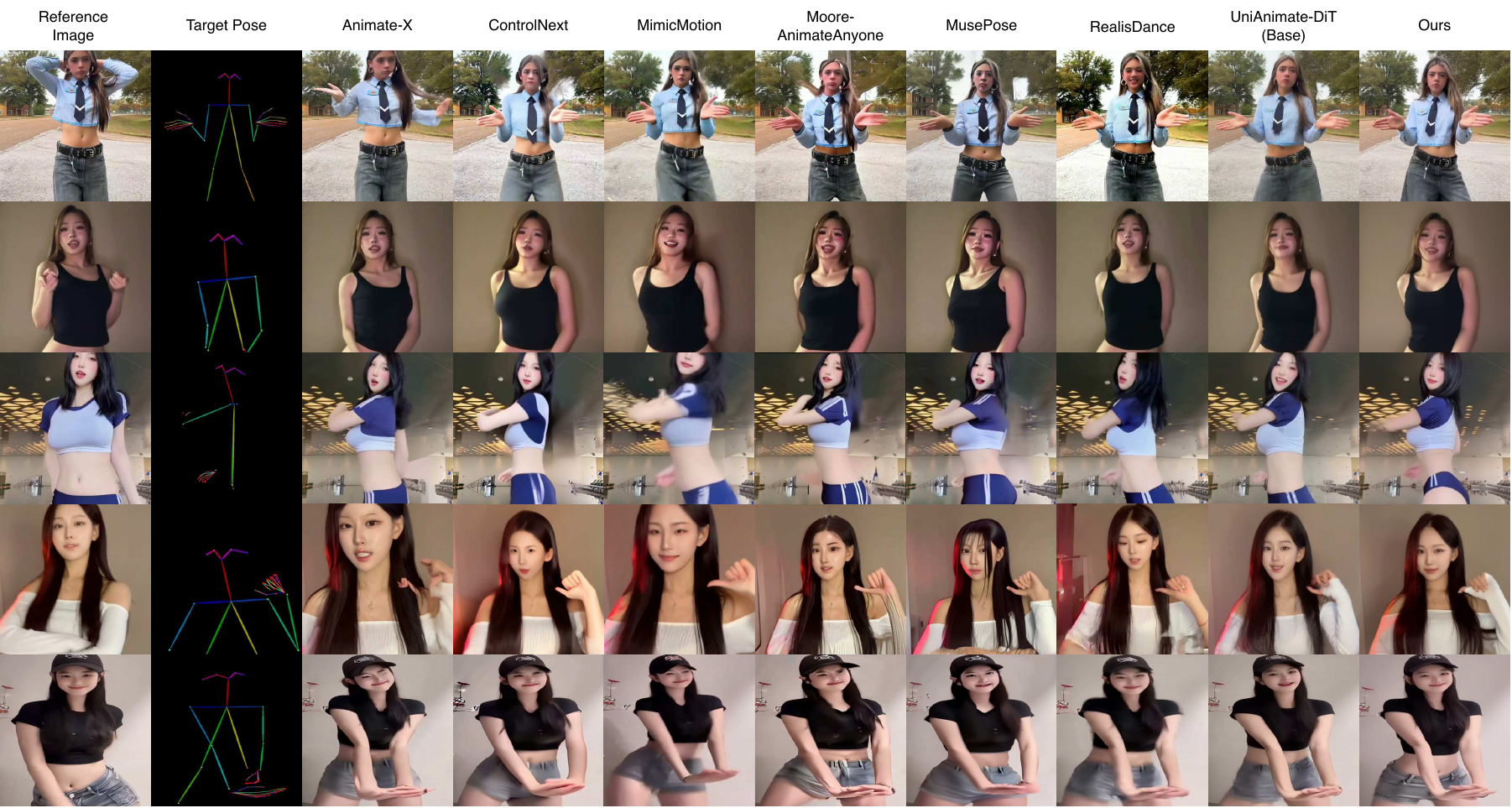}
  \caption{\textbf{In-Domain Comparisons: Diverse Appearance and Illumination.}
  Examples covering variations in skin tone, lighting conditions, and clothing
  texture. Our approach maintains consistent subject appearance and realistic
  fine-grained details across all scenarios, clearly outperforming state-of-the-art
  methods.}
  \label{fig:qual-supp2}
\end{figure*}

\begin{figure*}[p]
  \centering
  \includegraphics[width=\linewidth]{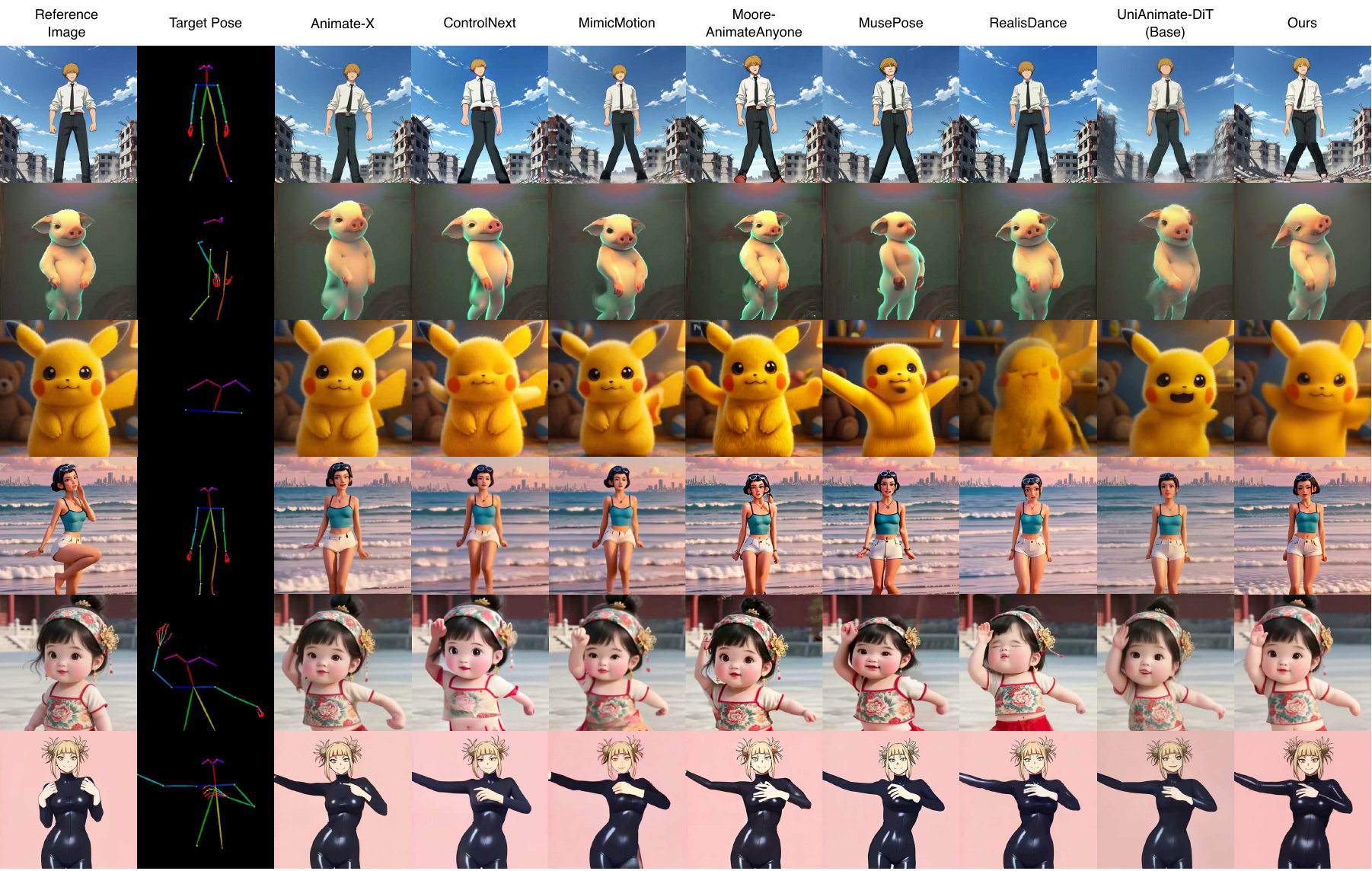}
  \caption{\textbf{Cross-Domain Comparisons: Animated Characters.}
  Examples from the Unseen100-Val evaluation set featuring 2D and 3D animated
  characters unseen during training. Our approach generalizes beyond photorealistic
  subjects, faithfully preserving stylized appearance details and character proportions
  while maintaining temporal coherence under complex motion.}
  \label{fig:qual-supp3}
\end{figure*}

We provide extended visual comparisons across subjects, motion types, and visual
domains, including cross-domain subjects unseen during training.
\Cref{fig:qual-supp1,fig:qual-supp2} give in-domain comparisons spanning diverse
motions, backgrounds, appearance, and illumination, and \cref{fig:qual-supp3} shows
cross-domain results on animated characters. Across all examples, the adapted model
preserves the subject's appearance while maintaining temporal coherence under
challenging lighting and complex motion.

\section{Test-Time Cost}
\label{app:cost}

This section reports the cost of serving one reference image on a single H100:
the wall-clock and peak GPU memory of every stage, how that memory is allocated, how it
amortizes over repeated use, and the implications for deployment.

\subsection{Measurement protocol}
\label{app:cost:protocol}
All numbers are measured on one NVIDIA H100 80\,GB (driver 580.173) on a single
TikTok test subject, with adaptation and inference at $832{\times}480$ and pseudo-frame
generation at the generator's native $576{\times}1024$, using the default recipe of
\cref{sec:exp:impl} ($K{=}5$ pseudo frames from $N{=}8$ reward-guided particles,
rank-$128$ adapter, $5$ adaptation steps). Peak memory is the maximum device
memory sampled every $250$\,ms for the lifetime of the stage, cross-checked against
the per-process figure. We report the device-level number because it includes the
CUDA context and allocator reserve and is therefore what a deployment must
provision. For the inference rows we also report the \texttt{torch} allocator peak: it runs $\sim\!1$\,GiB lower, and quoting only the allocator figure
would understate the requirement. Inference times exclude the one-off model load,
which a deployment pays once rather than per video.

\begin{table}[t]
  \centering
  \caption{\textbf{Per-stage test-time cost}, single H100 80\,GB, $832{\times}480$,
  one TikTok subject, with $81$-frame inference. \emph{Time} is generation only for
  the inference rows and wall-clock elsewhere. \emph{Peak GPU} is the maximum resident
  device memory during the stage, and \emph{Alloc.} the torch allocator peak where the
  pipeline reports it (protocol in \cref{app:cost:protocol}).}
  \label{tab:costbreakdown}
  \scriptsize
  \setlength{\tabcolsep}{4pt}
  \begin{adjustbox}{max width=\columnwidth}
\begin{tabular}{@{}lccc@{}}
    \toprule
    Stage & Time & Peak GPU (GiB) & Alloc.\ (GiB) \\
    \midrule
    \multicolumn{4}{@{}l}{\emph{One-time, per reference image}} \\
    \quad Pseudo-frame generation ($K{=}5$, $N{=}8$) & $8.7$\,min  & 41.7 & -- \\
    \quad ArcFace candidate selection                & $0.5$\,min  & 1.2  & -- \\
    \quad LoRA adaptation, $r{=}128$, $5$ steps      & $5.8$\,s  & 62.1 & -- \\
    \quad \textbf{Total} (time summed, memory maximum) & \textbf{$9.3$\,min} & \textbf{62.1} & -- \\
    \addlinespace[2pt]
    \multicolumn{4}{@{}l}{\emph{Per video}} \\
    \quad Inference, ours                            & $1.8$\,min  & 21.6 & 20.6 \\
    \quad Inference, base                            & $1.7$\,min  & 21.6 & 20.6 \\
    \bottomrule
  \end{tabular}
  \end{adjustbox}
\end{table}

\subsection{Composition of the memory peak}
\label{app:cost:memory}

\begin{table}[t]
  \centering
  \caption{\textbf{Composition of the adaptation memory peak.} Resident sizes
  follow from each checkpoint's parameter count and load precision: bf16 for the frozen
  modules, and fp32 with gradients and two AdamW moments for the low-rank update. The
  final line is the measured peak minus the resident modules above it.}
  \label{tab:memorybreakdown}
  \scriptsize
  \setlength{\tabcolsep}{4pt}
  \begin{adjustbox}{max width=\columnwidth}
\begin{tabular}{@{}lccc@{}}
    \toprule
    Component & Params & Resident (GiB) & Share \\
    \midrule
    \textbf{Frozen base model} & $\mathbf{23.0}$\textbf{B} & \textbf{43.6} & \textbf{70\%} \\
    \quad Video DiT $\theta$ (bf16)   & $16.4$B & 30.5 & 49\% \\
    \quad Text encoder umT5-xxl       & $5.3$B  & 10.6 & 17\% \\
    \quad Image encoder + VAE         & $1.3$B  & 2.5  & 4\%  \\
    \addlinespace[2pt]
    Update $\Delta\theta$ + grads + AdamW (fp32) & $0.61$B & 9.2 & 15\% \\
    Activations, workspace, allocator reserve    & --      & 9.4 & 15\% \\
    \midrule
    \textbf{Measured peak} &  & \textbf{62.1} &  \\
    \bottomrule
  \end{tabular}
  \end{adjustbox}
\end{table}

\Cref{tab:memorybreakdown} decomposes the adaptation peak. The frozen base model
accounts for $70\%$ of it, the video DiT contributing $49\%$ and the text encoder,
image encoder and VAE a further $21\%$ between them. The per-subject update accounts
for $15\%$, and is dominated not by the adapter weights ($613$M parameters,
$\approx\!4\%$ of the backbone) but by the optimizer state they require, since the
low-rank tensors are upcast to fp32 and AdamW keeps two moments per parameter. The
remaining $15\%$ is activations, workspace and allocator reserve.

The adaptation requirement is therefore a property of fine-tuning this animator
rather than of the supervision we propose. The Real-5 and reference-only
arms of \cref{tab:source} run the identical trainer at the identical rank and differ only in
which frames they consume, so they occupy the same footprint, as would any
adapter-based personalization of a backbone this size.

\noindent\textbf{Minimum GPU.} The minimum GPU is therefore set by the frozen
base model, not by our update: adaptation needs an $80$\,GB card, while inference
needs only $24$\,GB, the same requirement as running the base animator unadapted.

Inference requires far less memory despite using the same backbone, because it streams the
DiT: only a $6$B-parameter working set is resident and the remaining blocks are
fetched layer-by-layer during the forward pass.

\subsection{Deployment and amortization}
\label{app:cost:deploy}
The two costs are distinct. Adaptation is a one-time, server-side job of
$9.3$ minutes whose entire output is a $1.1$\,GiB adapter file. The $2.3$\,GiB
training checkpoint additionally carries optimizer state, which is not part of the
deployment payload. Once obtained, the adapter is reused for every video generated
from that reference. Inference is the recurring cost and is unchanged from the base
model, at identical peak memory and the same generation time to within run-to-run
variance, so a user never requires training-grade hardware: the adapter is downloaded and
inference runs at base cost.

\begin{table}[t]
  \centering
  \caption{\textbf{Amortization of the one-time cost.} Adaptation ($9.3$ min)
  is paid once per reference image, and each subsequent video of that reference costs
  only inference. Since adapted and base inference are the same computation
  (\cref{tab:costbreakdown}), we use a single render time for both, the measured
  $7.46$ min for a full-length $305$-frame video, so the ratio reflects the one-time
  cost alone. Average cost per video is $(9.3 + 7.46n)/n$, which decays as $1/n$ and
  is within $5\%$ of the unadapted base by $n{=}25$.}
  \label{tab:amortization}
  \scriptsize
  \setlength{\tabcolsep}{6pt}
  \begin{adjustbox}{max width=\columnwidth}
\begin{tabular}{@{}lccccc@{}}
    \toprule
    Videos from one reference $n$ & 1 & 5 & 10 & 25 & 100 \\
    \midrule
    Ours, min/video  & 16.8 & 9.3 & 8.4 & 7.8 & 7.6 \\
    Relative to base & $2.25\times$ & $1.25\times$ & $1.12\times$ &
                             $1.05\times$ & $1.01\times$ \\
    \bottomrule
  \end{tabular}
  \end{adjustbox}
\end{table}

\Cref{tab:amortization} quantifies the benefit of repeated use. Because
adaptation is paid per reference image and not per video, the average cost per video
falls as $1/n$: a reference animated $25$ times costs $1.05\times$ what the unadapted
base would cost for the same $25$ videos, and approaches parity beyond that. The
regime where the overhead dominates is the single-video case, where the one-time
stage is $1.25\times$ the inference it enables. Adaptation is thus better characterized as a provisioning cost, paid once when a reference is enrolled, than as a per-request
cost.

\section{Limitations}
\label{app:limit}
\begin{figure*}[!t]
  \centering
  \includegraphics[width=\linewidth]{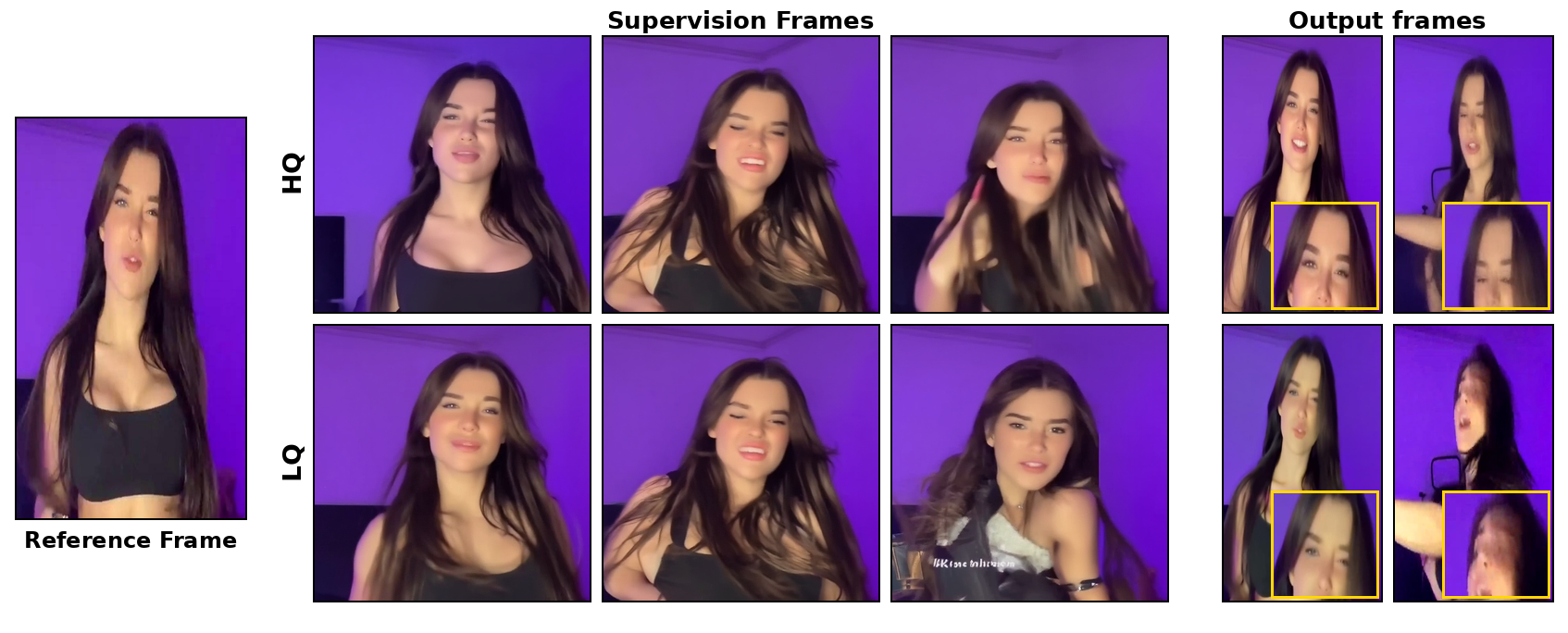}
  \caption{\textbf{Adaptation is bounded by pseudo-frame quality.}
  \emph{Top (HQ):} high-quality pseudo frames (\emph{Supervision}) yield faithful
  adapted outputs (\emph{Output frames}). \emph{Bottom (LQ):} degraded pseudo frames
  for the same subject yield weaker outputs. This is why the reference-only sampler
  selects high-quality frames before adaptation.}
  \label{fig:failure}
\end{figure*}

The method inherits the quality of its generator: when the generator produces weak
pseudo-ground-truth for a subject (for example under heavy occlusion, motion blur,
or strong out-of-domain appearance), the downstream per-subject adaptation is
correspondingly weaker, since adaptation can only bind the appearance that the
pseudo-ground-truth actually exposes. This motivates both the reference-only
selection that curates high-quality pseudo frames and the closed-loop direction
discussed in the conclusion, where the adapted model's own outputs re-supervise the
generator. A complementary safeguard is an automatic frame-quality filter that
rejects blurred or occluded pseudo frames before adaptation. \Cref{fig:failure}
illustrates how the adapted output tracks the quality of the pseudo-supervision.

\section{Ethical Considerations \& Broader Impact}
\label{app:ethics}
Our method lowers the cost of personalizing a human video model to a new subject:
it needs a single reference image, no subject video, and no manual labels. This
benefits creative content production, accessibility, and avatar applications, but
the same capability warrants a discussion of misuse.

\noindent\textbf{Risk of misuse.} The ease of synthesizing realistic video of a
person from one image raises the risk of non-consensual or deceptive media, such as
impersonation or misinformation. We condemn any use that depicts an individual
without their explicit consent, and we intend any release to be governed by terms
that prohibit such use.

\noindent\textbf{Data bias.} Like all generative models, ours can inherit biases in
its training data. If the underlying datasets underrepresent particular skin tones,
body types, ages, or styles of dress, personalization quality may be uneven across
groups. We encourage evaluation across diverse subjects, and our Unseen100-Val set
is curated to include a broad range of appearances and conditions toward this end.

\noindent\textbf{Mitigation strategies.} To navigate these risks responsibly, we
support provenance and detection measures, including digital watermarking of
generated video and research on deepfake-detection methods that distinguish
authentic from synthesized motion. Because our pipeline selects supervision by
reference identity, it is also compatible with consent-based safeguards that verify
the subject of the reference image before adaptation.

\end{document}